\documentclass[runningheads]{llncs}

\usepackage{eccv}

\usepackage{eccvabbrv}

\usepackage{graphicx}
\usepackage{bm}
\usepackage{booktabs}
\usepackage{multirow}
\usepackage{float}
\usepackage{makecell}

\usepackage{listings} 

\usepackage[accsupp]{axessibility}  

\usepackage{hyperref}

\usepackage{orcidlink}

\usepackage{subcaption}

\usepackage[table]{xcolor} 

\newcommand{\names}{CL4D }
\newcommand{\name}{CL4D}
\newcommand{\llms}{4DVLM }
\newcommand{\llm}{4DVLM}
\newcommand{\dataset}{DynAction4D}
\newcommand{\datasets}{DynAction4D }
\begin{document}


\title{\name: Contrastive Language–4D Pretraining for Vision-Language Reasoning in Dynamic Scenes}

\titlerunning{\names}

\author{Kumal Hewagamage$^\star$\inst{1}\orcidlink{0009-0000-5009-4621} \and
Isuranga Senavirathne$^\star$\inst{1}\orcidlink{0009-0006-8742-3699} \and
Sasika Amarasinghe$^\star$\inst{1}\orcidlink{0009-0007-6574-7948}
\and
Hasitha Gallella$^\star$\inst{1}\orcidlink{0009-0001-7597-7938}
\and
Dulanga Weerakoon$^\dagger$\inst{2}\orcidlink{0000-0002-4637-955X}
\and
Vigneshwaran Subbaraju$^\dagger$\inst{3}$^,$\inst{4}\orcidlink{0000-0002-4276-5939}
\and
Ranga Rodrigo$^\dagger$\inst{1}\orcidlink{0000-0002-1034-7513}}

\authorrunning{Hewagamage et al.}

\institute{University of Moratuwa, Sri Lanka
\email{\{hewagamagekln.21, senavirathneiub.21, amarasingheywsp.21, gallellammhhb.21,  ranga\}@uom.lk}
\and
Singapore-MIT Alliance for Research \& Technology (SMART) Centre, Singapore 
\email{dulanga.weerakoon@smart.mit.edu}\\
\and
Agency for Science, Technology and Research (A*STAR), Singapore\\
\and
CNRS@CREATE, Singapore\\
\email{Vigneshwaran\_Subbaraju@a-star.edu.sg}}

\maketitle

\footnotetext[1]{$^*$These authors contributed equally. $^\dagger$These authors jointly supervised this work.}

\footnotetext[2]{Accepted at ECCV 2026}

\begin{figure*}[!h]
    \centering
    \includegraphics[width=\textwidth]{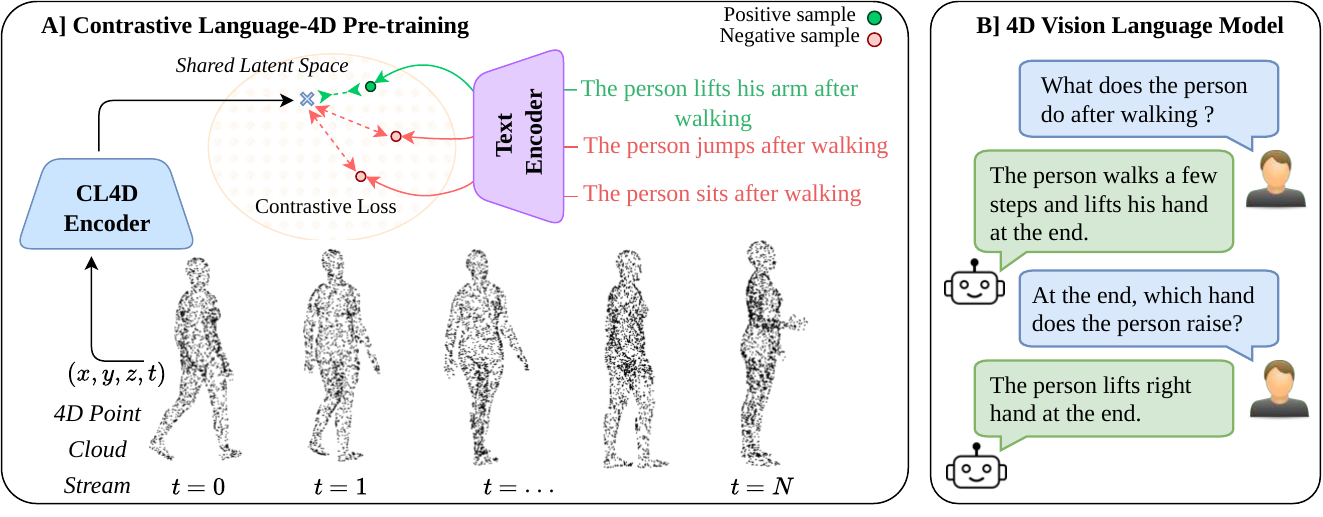} 
    \vspace{-0.25in}
    \caption{\textbf{Overview of the Language–4D framework:} We introduce CL4D, a contrastive pre-training paradigm that bridges the gap between 4D sequences and natural language. (A) Contrastive Language-4D Pre-training aligns 4D sequences with text in a shared latent space via a multi-modal contrastive loss. (B) 4D Vision-Language Model 
    which uses the learned representations for VLM downstream reasoning tasks.}
    \label{fig:teaser}
    \vspace{-0.4in}
\end{figure*}

\begin{abstract}

4D understanding and reasoning is a fundamental capability for embodied AI agents operating in dynamic physical environments. However, existing vision encoders are largely limited to static 2D images or 3D point clouds without temporal modeling, or to 2D videos that lack accurate geometric depth reasoning. Consequently, current approaches fail to jointly capture spatial structure and motion evolution in dynamic scenes. We present \name, the first foundational 4D vision encoder that directly operates on dynamic point clouds, trained with a contrastive learning objective to align spatio-temporal geometric representations with natural language descriptions. By learning a shared embedding space between text and 4D scene dynamics, \names enables zero-shot motion-to-text and text-to-motion retrieval in dynamic environments and serves as a foundational 4D vision encoder for downstream 4D vision--language tasks. Building on this encoder, we introduce \llm, a 4D vision--language model that conditions language generation on dynamic geometric representations. \llms is the first VLM designed to operate directly on 4D point clouds without relying on 2D images, 2D videos, or static 3D point clouds. We train \names and subsequently \llms on a newly constructed dataset termed \datasets\footnotetext[3]{Dataset and codes available at \url{https://4d-vision-uom.github.io/}.} capturing diverse human motions across varying object interactions and scene environments. Extensive experiments across multiple 4D human action benchmarks demonstrate that \names achieves state-of-the-art performance, with improvements of approximately $\sim$16.75\% over prior methods. Furthermore, \llms outperforms frontier video VLMs such as Gemini and GPT-5 even when these models are provided with RGB video sequences corresponding to the same scenes represented as 4D point clouds for \llm.

  \vspace{-0.2in}
\end{abstract}

\section{Introduction}  

The physical world is inherently four-dimensional, consisting of three spatial dimensions and one temporal dimension. Consequently, understanding and reasoning over dynamic 4D environments is a fundamental capability for embodied artificial intelligence (EAI). Models designed for such settings must develop a unified representation that captures both geometric structure and temporal dynamics within the spatio-temporal space of the physical world.


Existing approaches for physical world perception and reasoning rely predominantly on Vision Language Models (VLMs). However, most VLMs are designed for static 2D imagery, extracting visual features from single images, as in Llama \cite{touvron2023llama}, or extending to video-based representations, as in VideoLLM \cite{chen2023videollm}. While video VLMs incorporate temporal signals, they remain fundamentally grounded in 2D pixel observations and lack explicit 3D geometric reasoning. More recent works, such as Ges3ViG \cite{mane2025ges3vig} and PointLLM \cite{xu2024pointllm}, extend multi-modal reasoning to 3D point cloud representations captured via LiDAR. Nevertheless, these methods are restricted to static 3D scenes and do not model temporal dynamics, limiting their applicability in dynamic physical environments. 

Developing effective 4D reasoning models requires, as a foundational step, a robust 4D vision encoder. In 2D, 3D, and 2D with temporal modeling settings, contrastive pretraining \cite{radford2021clip} has emerged as the dominant paradigm for learning vision encoders aligned with language representations. Such encoders are trained to project visual and textual features into a shared embedding space, enabling strong transfer to downstream vision language tasks including Visual Question Answering (VQA) \cite{antol2015vqa}, Visual Grounding (VG) \cite{deng2021transvg_visual_grounding}, and image or video captioning \cite{abdar2024_video_captioning}. Despite the success of contrastive alignment in these domains, no prior work has extended this paradigm to 4D representations that capture dynamic 3D scenes. In particular, existing approaches do not learn language aligned embeddings from temporally evolving point clouds that jointly encode both geometry and motion. 
To address this gap, we introduce \emph{\textbf{C}}ontrastive \textbf{\emph{L}}anguage-\textbf{\emph{4D}} (\emph{\textbf{\name}}), the first foundational 4D vision encoder. \names is trained using a contrastive learning objective to align dynamic 4D point cloud sequences representing human actions with their corresponding natural language descriptions. By learning a shared embedding space between spatio-temporal geometric features and action semantics, \names enables zero-shot action-to-text retrieval and establishes a foundation for language-grounded reasoning over dynamic 4D environments.


Building upon the proposed \names 4D vision encoder, we further introduce \emph{\textbf{4D}} \textbf{\emph{V}}ision \textbf{\emph{L}}anguage \emph{\textbf{M}}odel (\emph{\textbf{\llm}}), the first VLM designed to directly reason over dynamic 4D point clouds. Unlike prior VLMs that operate on 2D images, videos, or static 3D scenes, \llms is grounded in temporally evolving 3D scenes, enabling joint spatial and temporal reasoning in dynamic physical environments.


In summary, we make the following \textbf{key contributions}:
\vspace{-0.05in}
\begin{enumerate}

\item \textbf{\name: A Contrastively Aligned 4D Vision Encoder.}  
We introduce \name, the first foundational 4D vision encoder trained to align dynamic 4D point cloud sequences with natural language using a contrastive learning objective. Unlike prior 4D encoders that rely on classification over a fixed set of action labels, \names learns a shared embedding space between spatio-temporal scene dynamics and textual descriptions. This enables zero-shot cross-modal retrieval between motion and text in dynamic environments and establishes a foundational representation for 4D vision–language tasks. Empirically, \names improves Recall@1 for text–motion retrieval by up to \textbf{16.75\%} compared to prior 4D encoders.

\vspace{0.08in}

\item \textbf{\llm: The First 4D Vision–Language Model.}  
Building upon \name, we propose \llm, the first Vision–Language Model that directly reasons over temporally evolving 4D point clouds without relying on RGB images, videos, or static 3D scenes. By conditioning language generation on geometry-aware spatio-temporal embeddings, \llms enables language-grounded reasoning in dynamic physical environments. Remarkably, \llms achieves \textbf{state-of-the-art performance even compared to frontier video-based VLMs such as Gemini and GPT-5}, despite those models operating on RGB video inputs.

\vspace{0.08in}

\item \textbf{\dataset: A Benchmark for Language–4D Alignment.}  
To enable systematic evaluation and large-scale contrastive training of 4D vision models, we introduce \dataset, a benchmark dataset consisting of dynamic human actions represented as temporally evolving 4D point cloud sequences paired with natural language descriptions. The dataset spans diverse action categories and scene environments with varying object layouts, providing a structured benchmark for studying language–4D alignment and evaluating 4D VLMs.

\end{enumerate}

\section{Related Work}

\subsubsection{Contrastive Pretraining for Vision--Language Alignment:}

Early vision encoders were largely developed using supervised learning on large-scale datasets such as ImageNet \cite{deng2009imagenet}, leading to architectures including AlexNet \cite{krizhevsky2012imagenet}, VGGNet \cite{simonyan2014very}, and ResNet \cite{he2016deep}. These models were trained using cross-entropy classification objectives and later reused as general visual feature extractors for downstream tasks. However, such representations are tied to predefined label spaces and are not inherently aligned with language.

Contrastive pretraining addressed this limitation by aligning visual and textual representations in a shared embedding space. CLIP \cite{radford2021clip} introduced large-scale language–image contrastive learning, enabling strong zero-shot transfer across tasks. Subsequent works extended this paradigm to video and 3D representations. For videos, methods such as VideoCLIP \cite{xu2021videoclip}, CLIP4Clip \cite{luo2022clip4clip}, and ActionCLIP \cite{wang2023actionclip} learn joint video–text embeddings by modeling temporal dynamics in RGB frame sequences. In the 3D domain, approaches including PointCLIP \cite{zhang2022pointclip}, ULIP \cite{xue2023ulip}, and OpenShape \cite{liu2023openshape} align static point clouds with language through multi-view projections or joint multimodal pretraining. 

Despite these advances, existing methods largely operate on static images, RGB video frames, or static 3D geometry, and do not explicitly model temporally evolving 3D environments.

\subsubsection{4D Vision Encoders:}

Modeling dynamic point cloud sequences introduces additional challenges due to the irregular and unordered nature of point clouds across time. Early approaches focused on extracting spatio-temporal features for action recognition rather than language alignment. For example, Motion PointNet \cite{huang2024_motion_pointnet} models temporal variations using a PointNet-style architecture with temporal aggregation, optimized using cross-entropy objectives for closed-set action classification.

More recent works attempt to bridge motion understanding and language by learning cross-modal representations. Motion Patches \cite{yu2024exploring} converts human motion sequences into grid-like patches compatible with Vision Transformers pretrained on images, enabling contrastive alignment between motion and language representations. While these approaches demonstrate promising zero-shot capabilities, they rely on structured skeleton representations rather than raw dynamic point clouds.

Consequently, existing methods either rely on classification-based objectives or structured motion representations, limiting their applicability for open-vocabulary reasoning over raw dynamic 4D environments.

\subsubsection{Vision Language Models:}

Large vision–language models have recently achieved strong performance on multimodal reasoning tasks such as visual question answering, visual grounding, and captioning. Systems such as GPT-5 \cite{gpt5}, Gemini \cite{geminiteam2025geminifamilyhighlycapable}, and LLaVA \cite{liu2023visual} integrate visual encoders with large language models to enable reasoning over images and videos. Dedicated video VLMs such as VideoLLM \cite{chen2023videollm} and Video-LLaVA \cite{lin2024video} further explore spatio-temporal reasoning using RGB video inputs.

Recent works have also explored grounding language models in 3D scenes. Approaches such as M3DRefCLIP \cite{zhang2023multi3drefer} and Ges3ViG \cite{mane2025ges3vig} study 3D visual grounding, while PointLLM \cite{xu2024pointllm}, LL3DA \cite{chen2024ll3da}, and OneLLM \cite{han2024onellm} extend instruction tuning to static point clouds.

However, existing methods operate either on 2D RGB data or static 3D geometry. None directly support language-based reasoning over dynamic 4D point cloud sequences. In contrast, \llms enables vision–language reasoning directly on temporally evolving 4D point clouds, allowing joint understanding of geometry and motion dynamics.
\section{Methodology}
Our framework is designed to enable language-grounded reasoning over dynamic 4D point cloud sequences. We adopt a two-stage training strategy Fig.~\ref{fig:methodology}. In the first stage, we learn semantically aligned spatio-temporal representations through contrastive Language–4D pretraining (\name). In the second stage, we rely on the pretrained 4D encoder to construct a 4D Vision–Language Model (\llm) that supports instruction following and visual question answering over dynamic point clouds. To facilitate this two-stage learning process, we introduce the \datasets{}dataset.

\subsection{\datasets Dataset}
\label{sec:4d_Data}


We first present our proposed benchmark dataset for Language–4D pretraining aimed at enabling vision–language reasoning in dynamic physical environments. The proposed dataset primarily consists of dynamic 4D point cloud sequences capturing various human actions across diverse scene environments, along with their corresponding textual descriptions. \datasets comprises four primary segments, each representing distinct dynamic scenarios. We now describe each dataset segment of the \datasets dataset.


\begin{figure}[!h]
    \centering
    \includegraphics[width=1\linewidth]{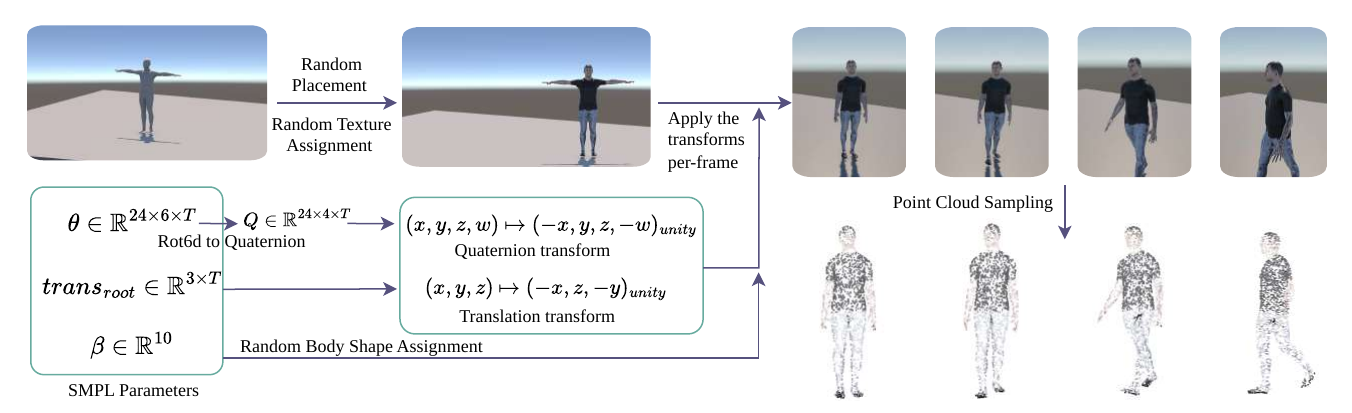}
    \caption{\textbf{Overview of the \dataset{} generation pipeline.} SMPL parameters 
are first converted to Unity-compatible animations, then augmented 
using various methods such as random body shape assignment and 
random texture variations from \cite{varol2017learning}. The resulting meshes are then uniformly 
sampled in Unity to generate human action point cloud sequences. This pipeline yields highly diverse, temporally evolving point cloud sequences suitable for training complex dynamic scene understanding models.
}
    \label{fig:data_generation_pipeline}
\end{figure}



\begin{figure}[!h]
    \centering
    \begin{subfigure}{0.49\linewidth}
        \vspace{0pt}
        \centering
        \includegraphics[width=\linewidth]{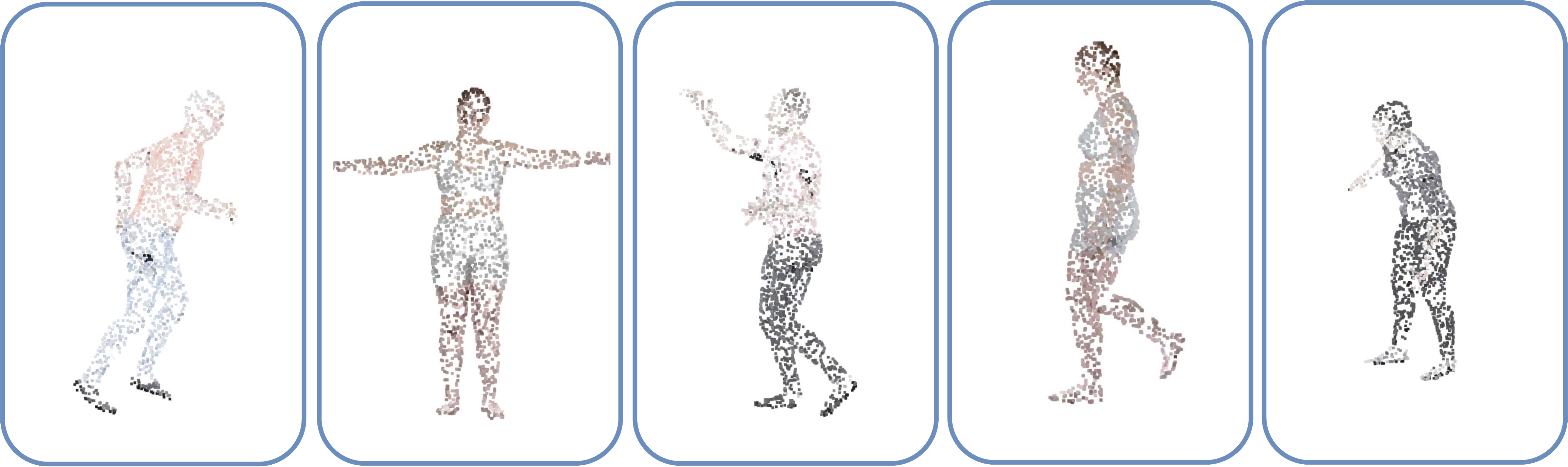}
        \caption{\textbf{\dataset-HumanOnly}}
        \label{fig:humanml_dataset_samples}
    \end{subfigure}
    \hfill
    \begin{subfigure}{0.49\linewidth}
        \centering
        \vspace{0pt}
        \includegraphics[width=\linewidth]{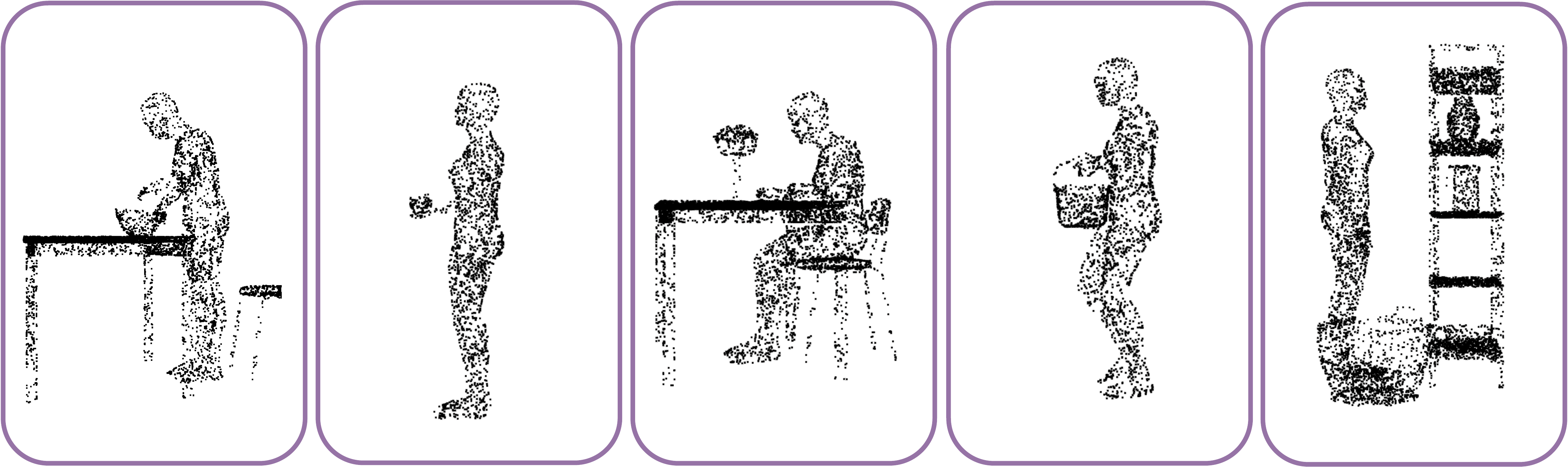}
        \caption{\textbf{\dataset-ObjInteractions}}
        \label{fig:humoto_dataset}
    \end{subfigure}

    \vspace{1em}

    \begin{subfigure}{0.96\linewidth}
        \centering
        \includegraphics[width=\linewidth]{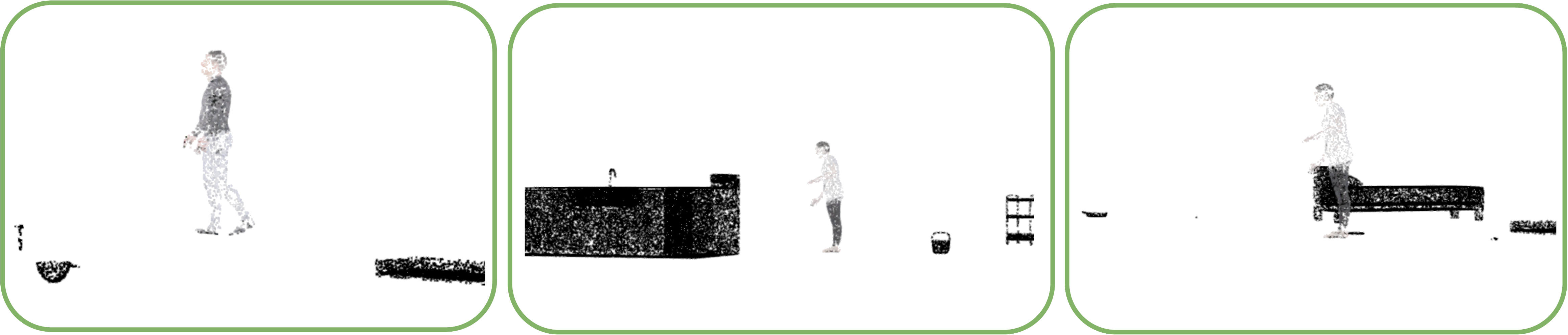}
        \caption{\textbf{\dataset-Cluttered}}
        \label{fig:cluttered_humanML_dataset_samples}
    \end{subfigure}

    \caption{\textbf{Sample frames from point cloud sequences across the three dataset segments in \dataset.}
  \textbf{(a) HumanOnly:} Five sequences, each depicting a distinct human body shape and action.
  \textbf{(b) ObjInteractions:} Five sequences, each featuring a human interacting with a different object.
  \textbf{(c) Cluttered:} Three sequences showing humans performing actions in cluttered environments.
}
\vspace{-0.2in}
\end{figure}

\paragraph{\textbf{1. HumanOnly Segment}}

As shown in Figure~\ref{fig:humanml_dataset_samples}, HumanOnly  segment, considers a variety of actions performed by a single human, derived from the HumanML3D dataset~\cite{humanml3d}. HumanML3D is a human motion–language dataset that covers a broad range of activities, including daily actions (e.g., walking, jumping), sports (e.g., swimming, playing golf), acrobatics (e.g., cartwheel), and artistic movements (e.g., dancing). This dataset provides SMPL parameters for each motion sequence along with corresponding natural language descriptions that explain the actions. The training split for HumanOnly contains 23k sequences, while the test split consists of 4k sequences.

\paragraph{\textbf{2. ObjInteractions Segment:}}

This segment considers human actions that involve interactions with objects. To support this, we construct 4D scenes using human–object interaction sequences from the Humoto dataset \cite{lu2025humoto}, which models diverse daily activities involving 72 objects comprising 735 distinct object interactions.The training split for  ObjInteractions contains 510 sequences, while the test split consists of 219 sequences.


\paragraph{\textbf{3. Cluttered Segment:}}

This Cluttered segment considers a scenario where a human performs various actions in cluttered scene environments. To construct such a scenario, we place human action sequences taken from the HumanML3D dataset within scenes populated with Humoto objects at varying placements and clutter conditions. The training split for the Cluttered segment contains 23k sequences, while the test split consists of 4k sequences.




\subsubsection{Textual Data Generation}

Text descriptions are inherited from the original action datasets, where each motion sequence is paired with a natural language annotation. These text–motion pairs are used for language-aligned contrastive pre-training.

\paragraph{\textbf{ 4D-VQA Segment:}}
For training the proposed \llm,
we create a novel 4D-VQA dataset. For this we employ videos rendered in Unity from existing human action sequences generated from the HumanML3D dataset.
To synthesize this dataset segment, we prompt Gemini 3.0 Flash \cite{geminiteam2025geminifamilyhighlycapable} with rendered Unity videos and HumanML3D text descriptions, generating structured queries designed to test spatial and temporal reasoning. The QA pairs span three distinct motion-analysis categories: (a) \textbf{Action:} Focuses on high-level semantic understanding and sequence classification. (b) \textbf{Body-Spatial Qualities:} Focuses on 3D spatial reasoning, geometric relationships, and limb positioning. (c) \textbf{Temporal Qualities:} Focuses on the dynamics and frame-to-frame state changes of the motion.




\subsection{\name: Contrastive Language-4D Pre-training}
\label{sec:cl4d_pretraining}

\begin{figure*}[t]
    \centering
    \includegraphics[width=1.06\textwidth]{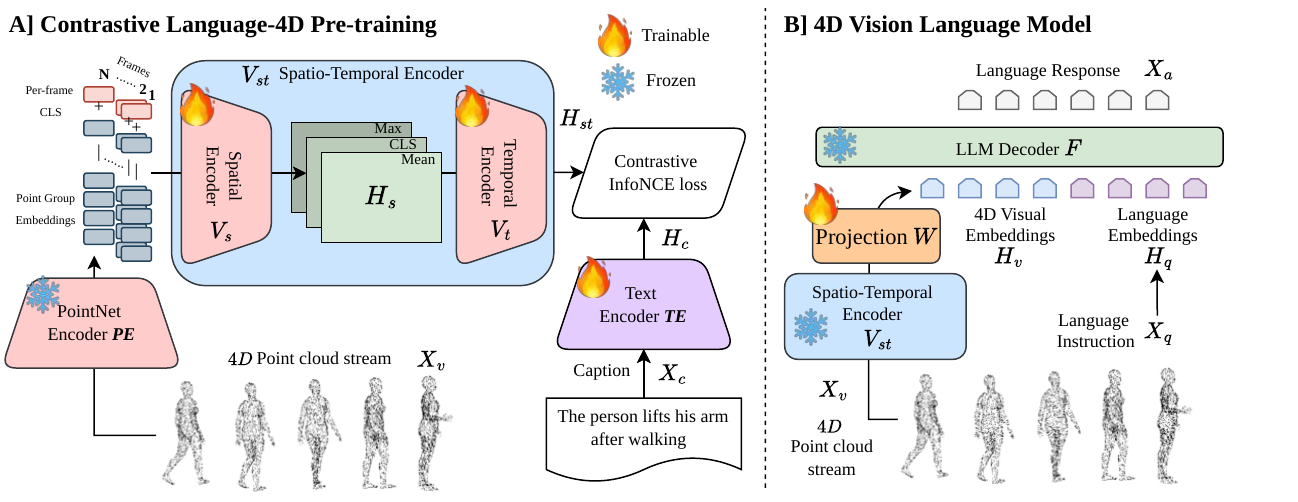} 
    \caption{
    \textbf{\names framework overview:}
    (A) \textbf{Contrastive Language-4D Pre-training} (Sec.~3.1): 
    A dynamic point cloud sequence $(x,y,z,t)$ is encoded by the PointNet-based ${PE}$ and a spatio-temporal encoder ${V}_{st}({V}_{s}, {V}_{t})$ to obtain 4D visual embeddings, which are aligned with text features from ${TE}$ via a contrastive objective.
    (B) \textbf{4D Visual Question Answering} (Sec.~3.2): 
    The learned ${V}_{st}$ representations are fused with language embeddings in a LLaVA setup to enable reasoning over dynamic point clouds and generate action-aware responses.
    }
    \label{fig:methodology}
    \vspace{-0.2in}
\end{figure*}

In the first stage of training, we introduce a novel Spatio-temporal 4D Vision Encoder as illustrated in Fig.~\ref{fig:methodology}(A). This encoder learn semantically aligned spatio-temporal representations via contrastive language-4D pre-training (CL4D). Our objective is to learn language-aligned representations for dynamic point cloud sequences $(x,y,z,t)$ through large-scale contrastive pre-training.

\paragraph{\textbf{Point Encoder (PE):}}

Consider a 4D sequence with $N$ point cloud frames. Each frame of the input sequence is first processed by a standard PointNet~\cite{pointnet} encoder ${PE}$ that operates on local point groups of each frame. We divide each frame into $k$ number of groups, each containing $g$ number of points. The encoder produces $k$ embeddings of dimension $d_1$ for each frame $p_t \in X_v$ point cloud stream. We prepend a learnable ${CLS}$ token, resulting in $(k+1)$ embeddings per frame that form the feature tensor $f_t \in \mathbb{R}^{(k+1) \times d_1}$ for the input to the spatial encoder $V_s$, as defined in Eq.~\eqref{eq:pointnet_embedding}.


\paragraph{\textbf{Spatio-Temporal Transformer Encoder ($V_{st}$):}}
To jointly model the spatial structure and temporal dynamics of these $f_t$ features from $PE$, we introduce a transformer based encoder defined as $V_{st}$, which is a combination of the spatial encoder $V_s$ and the temporal encoder $V_t$. The processing order of this module is defined in Eqs.~\eqref{eq:spatial}--\eqref{eq:temporal}.


\vspace{-0.1in}

\begin{equation}
f_t = \mathrm{PE}(p_t) \in \mathbb{R}^{(k+1) \times d_1}, \quad 
p_t = X_v[t], \quad t = 1,\dots,N
\label{eq:pointnet_embedding}
\vspace{-0.2in}
\end{equation}

\vspace{-0.1in}

\begin{equation}
\tilde{h}_t = \{ \mathrm{MAX}(h_t), \mathrm{CLS}(h_t), \mathrm{MEAN}(h_t) \} \in \mathbb{R}^{3 \times d_2}, \quad h_t = V_s(f_t) \in \mathbb{R}^{d_2}
\label{eq:spatial}
\vspace{-0.1in}
\end{equation}

\vspace{-0.1in}

\begin{equation}
H_s = \{ \tilde{h}_1, \tilde{h}_2, \dots, \tilde{h}_N \} \in \mathbb{R}^{3 \times d_2 \times N} \quad for \quad t = 1,\dots,N
\label{eq:stack}
\end{equation}

\vspace{-0.3in}

\begin{equation}
 H_{st} = V_t(H_s) \in \mathbb{R}^{d_3}
 \label{eq:temporal}
\end{equation}


The ${V}_{s}$ captures global spatial relationships extracted to $f_t$ for each frame by ${PE}$ and ${V}_{t}$ models the inter-frame temporal dependencies across all $N$ frames. The embeddings inside $V_{st}$ transformer blocks are $d_3$ dimensional.



The input $f_t$ tensors will go through $V_s$ individually and produce the $h_t$ sequence of embeddings. From each $h_t$ we compute means, max, and cls embeddings and stack up for a $\tilde{h}_t$ per each frame as in Eq.~\eqref{eq:spatial}. After processing all the $N$ frames, we collect all the $\tilde{h}_t$ in to a single $3\times d_2 \times N$ dimensional 3D tensor $H_s$ as defined in Eq.~\eqref{eq:stack}. Then the $H_s$ 3D tensor will be patched up to $m \times m$ patches to follow similar to a standard vision transformer. It will create a new $(d_2 \times N)/(m \times m)$ length patch sequence. These patches will be encoded again into the $d_3$ dimensional embeddings through a linear projection and fed to the $V_t$ transformer blocks to model temporal relationships. At the end of $V_t$ we get the mean embedding as $H_{st}$ of the output embedding sequence of the transformer (Defined in Eq.~\eqref{eq:temporal}).

In our best setup, we used embedding dimensions $d_1 = 512$, $d_2=512$ and $d_3 = 768$. Point Encoder $PE$ produces with $k=64$ embeddings per frame with $g=32$ group size for the \dataset{} dataset. Both ${V}_{s}$ and ${V}_{t}$ transformers have 12 Attention Heads and MLP Ratio is 4 while transformer depth is ${V}_{s}$ = 4 and ${V}_{t}$ = 12 for each. ${V}_{s}$ is initialized from scratch and temporal encoder ${V}_{t}$ is initialized from $ViT-B/16$ weights prior to training, enabling stable optimization and improved motion modeling. ${V}_{t}$ used with patch size $m$ = 16 to process the $H_s \in \mathbb{R}^{3 \times 512 \times 32}$  which created $32 \times 2$ patches.




\paragraph{\textbf{Text Encoder (TE):}}
Language descriptions ($X_c$) are encoded using $distilbert-base-uncased$ ~\cite{sanh2019distilbert}. We use the CLS token representation ($H_c$) as the global text embedding which has $d_3$ dimensions. 

\paragraph{\textbf{Contrastive Alignment Objective:}}
Let $\{v_i\}_{i=1}^B$ denote visual embeddings ($H_{st}$) obtained by mean pooling the output token sequences of $\mathrm{V}_{st}$ for the batch size $B$, and $\{t_i\}_{i=1}^B$ denote the corresponding CLS embeddings ($H_c$) from the text encoder. 
We optimize a symmetric cross-entropy objective over the $B \times B$ similarity matrix using cosine similarity $s(\cdot,\cdot)$ with temperature $\tau$:

\begin{equation}
\mathcal{L}_{m2t} = 
-\frac{1}{B} \sum_{i=1}^{B}
\log 
\frac{\exp(s(v_i,t_i)/\tau)}
{\sum_{j=1}^{B} \exp(s(v_i,t_j)/\tau)},
\end{equation}

\begin{equation}
\mathcal{L}_{t2m} = 
-\frac{1}{B} \sum_{i=1}^{B}
\log 
\frac{\exp(s(v_i,t_i)/\tau)}
{\sum_{j=1}^{B} \exp(s(v_j,t_i)/\tau)},
\end{equation}

\begin{equation}
\mathcal{L} = \mathcal{L}_{m2t} + \mathcal{L}_{t2m}.
\end{equation}

This objective encourages aligned visual-text pairs to have high similarity while pushing apart mismatched pairs within the batch $B$. This alignment enables semantically grounded 4D representations that jointly encode geometric structure and motion dynamics. Furthermore, \names can serve as a general 4D vision backbone for a range of 4D vision-language reasoning tasks, where language and dynamic point clouds are projected into a unified embedding space. 



\paragraph{\textbf{Training Details:}}
Using the proposed contrastive alignment objective, \names is trained for 150 epochs with a batch size of $B$ = 72. Model parameters are optimized using AdamW optimizer with decoupled learning rates: $1.0 \times 10^{-4}$ for the $V_{st}$ components and projection heads, and $1.0 \times 10^{-5}$ for the text encoder.



\subsection{\llm: 4D Vision Language Model}
\label{sec:4d_vlm}

Building upon the contrastive pre-trained 4D vision encoder, we construct a 4D Vision-Language Model (Fig.~\ref{fig:methodology}(B)) capable of performing vision–language reasoning directly on 4D point clouds.

\paragraph{\textbf{Visual Token Projection:}}

We first take the spatio-temporal embeddings produced by $V_{st}$, and linearly project them to a 4096-dimensional space to match the hidden dimension of LLaMA.


\paragraph{\textbf{Language Backbone:}}
We initialize the language model from Llava-v1.5-7b, which is based on the Vicuna 7B backbone. The projected visual tokens are injected into the language model following the standard LLaVA-style multi-modal conditioning mechanism \cite{liu2023visual}.


\paragraph{\textbf{Training Objective:}}
The model is optimized using the standard auto-regressive language modeling loss employed in LLaVA. Given a visual sequence and a question $X_q$, the model generates an answer conditioned on projected 4D tokens:

\begin{equation}
\mathcal{L}_{\text{VLM}} = 
-\sum_{t} \log p(y_t \mid y_{<t}, X_q, \mathrm{Proj}(\mathrm{V}_{st})).
\vspace{-0.2in}
\end{equation}

\paragraph{\textbf{Training Setup:}}

With the proposed training objective, we freeze the language encoder and the \names spatio-temporal 4D encoder, and train only the projection layer for 10 epochs using the VQA split of \datasets. Training is performed with  a learning rate of $1.0 \times 10^{-3}$ and the AdamW optimizer.



\section{Results}

\subsection{Evaluation of \names}

\vspace{-0.1in}
\begin{table*}[!h]
\vspace{-0.2in}
\centering
\caption{ \textbf{Motion-text retrieval results on \datasets{}segments and RH20T dataset.} The results show the Recall@1  for both Batch and Global settings under text-to-motion and motion-to-text retrieval using training batch size = 72 and test batch size = 32. \names and \name-mini outperforms the other 4D encoders in all the \dataset{} segments and RH20T dataset.}
\label{tab:cl4d_vs_others}
\resizebox{\textwidth}{!}{
\begin{tabular}{llcccc}
\toprule
\multirow{2}{*}{\makecell{Dataset}} & \multirow{2}{*}{Method} 
& \multicolumn{2}{c}{Text-motion retrieval} 
& \multicolumn{2}{c}{Motion-text retrieval} \\
\cmidrule(lr){3-4} \cmidrule(lr){5-6}
& & \makecell{R@1$\uparrow$ \\ (Batch)} & \makecell{R@1$\uparrow$ \\ (Global)} 
& \makecell{R@1$\uparrow$ \\ (Batch)} & \makecell{R@1$\uparrow$ \\ (Global)} \\


\midrule
\multirow{4}{*}{\makecell{\datasets \\ HumanOnly}}
& P4Transformer  \cite{p4transformer}  & 53.57 & 3.66 & 58.05 & 5.33 \\
& PST-Transformer \cite{pst_transformer}  & 49.09 & 2.69 & 51.73 & 3.34 \\
& Motion PointNet  \cite{huang2024_motion_pointnet}   & 51.91 & 2.34 & 55.78 & 3.13 \\
& \textbf{\names (Ours)}             & \textbf{70.32} & \textbf{8.07} & \textbf{68.62} & \textbf{8.02} \\


\midrule
\multirow{4}{*}{\makecell{\datasets \\ ObjInteractions}}
& P4Transformer  \cite{p4transformer}  & 26.55 & 5.94 & 24.93 & 7.31 \\
& PST-Transformer \cite{pst_transformer}  & 30.37 & 8.22 & 25.50 & 7.31 \\
& Motion PointNet  \cite{huang2024_motion_pointnet}   & 41.37 & 12.33 & 36.18 & 11.87 \\
& \textbf{\names (Ours)}             & \textbf{49.40} & \textbf{23.29} & \textbf{46.97} & \textbf{22.37} \\

\midrule
\multirow{4}{*}{\makecell{\datasets \\ Cluttered}}
& PST-Transformer \cite{pst_transformer} & 30.86 & 0.90 & 33.29 & 0.93 \\
& Motion PointNet \cite{huang2024_motion_pointnet}   & 41.71 & 1.23 & 43.24 & 1.53 \\
& \textbf{\names (Ours)}             & \textbf{55.07} & \textbf{3.11} & \textbf{51.94} & \textbf{2.71} \\

\midrule
\midrule
\multirow{4}{*}{\makecell{RH20T \cite{fang2024rh20t}}}
& P4Transformer  \cite{p4transformer}  & 9.51 & 0.84 & 5.77 & 0.84 \\
& PST-Transformer \cite{pst_transformer}  & 54.76 & 20.17 & 39.61 & 15.13 \\
& Motion PointNet  \cite{huang2024_motion_pointnet}   & 68.21 & 31.09 & 52.28 & 24.37 \\
& \textbf{\names Mini (Ours)}             & \textbf{77.24} & \textbf{37.82} & \textbf{57.10} & \textbf{27.73} \\

\bottomrule
\end{tabular}
}
\vspace{-0.3in}
\end{table*}

\subsubsection{Evaluation Setup:}
We evaluate \names on two primary retrieval tasks. (a) \textit{Text-to-motion retrieval}, where a given text embedding is matched against a batch of 4D scene embeddings produced by \names, with only one embedding corresponding to the correct match. (b) \textit{Motion-to-text retrieval}, where a given 4D scene embedding obtained from \names is matched against a batch of textual embeddings, with only one text embedding corresponding to the correct scene.

We adopt Recall@$k$ (R@$k$)~\cite{petrovich2023tmr} as the primary evaluation metric. In the \textit{R@$k$ (Batch)} setting, retrieval is performed within a randomly sampled batch of embeddings. In contrast, \textit{R@$k$ (Global)} evaluates retrieval across the entire test dataset, where the query must retrieve the correct match from all available embeddings. \datasets is the primary benchmark for evaluation of \name. 


\subsubsection{Comparison with Prior 4D Encoders:}

\names represents the first 4D encoder trained with a contrastive learning objective that aligns scene and text embeddings within a unified representation space. In contrast, prior 4D encoders are typically trained using a standard cross-entropy objective to classify motion embeddings into a predefined and finite set of action labels. As a result, such encoders are not naturally suited for cross-modal retrieval tasks such as text-to-motion or motion-to-text retrieval when given a 4D point cloud as input.

To evaluate the effectiveness of \names against prior 4D encoders, we re-adapt these methods using the same contrastive alignment objective and train them on the proposed \datasets dataset. The comparison results are summarized in Table~\ref{tab:cl4d_vs_others}. Notably, \names achieves improvements of 16.75\%, 8.1\%, and 13.36\% in R@1 under the batch setting for text-to-motion retrieval on the HumanOnly, ObjInteractions, and Cluttered segments of \dataset, respectively.

These results demonstrate that \names enables stronger zero-shot alignment between textual descriptions and dynamic 4D scene embeddings without relying on a predetermined and finite set of action labels, as required by prior classification-based approaches.

\subsubsection{Generalizability of \names:}

To further demonstrate the generalizability of \names to real-world point cloud data involving actions performed by non-human agents, we evaluate \names on the robotic manipulator action dataset RH20T \cite{fang2024rh20t} in Table \ref{tab:cl4d_vs_others}. RH20T comprises of manipulation sequences across diverse skills, contexts, robots, and camera viewpoints, all collected in the real world. \name-mini achieves state-of-the-art performance, surpassing all baseline methods by at least 9.03\% in the batch setting for text-to-motion retrieval and 4.82\% for motion-to-text retrieval for tested splits.

\subsubsection{Ablation \& Sensitivity Studies:}

In Table~\ref{tab:cl4d_ablations}, we evaluate several architectural variations of \names. In particular, we vary the ViT backbone used in the temporal encoder of \names among the Tiny, Base, Small, and Large variants. Among these configurations, \names with the ViT-Small backbone achieves the highest recall accuracy for cross-modal retrieval under the batch evaluation setting.

When the text encoder is frozen (\textit{w. Frozen Text Enc.}), retrieval accuracy decreases across all metrics, highlighting the importance of jointly optimizing both the 4D scene and textual embeddings for stronger contrastive alignment in the shared embedding space. When SigLIP loss is used (\textit{w. SigLIP loss}) instead of the standard InfoNCE loss, performance across all metrics also decreases.

Across all ViT variants, we initialize the temporal encoder with pretrained ViT weights. In the \textit{w. Untrained ViT} setting, we instead use an untrained ViT-Base model for the temporal encoder. This configuration leads to lower accuracy, indicating the importance of prior 2D-learned ViT features for effective temporal modeling. The CL4D-mini, a parameter-efficient version of CL4D yields only a marginal drop in Recall@1 while achieving an approximate 60\% reduction in GFLOPs compared to CL4D with ViT-Tiny. CL4D-mini employs a reduced spatial encoder along with ViT-Tiny. Finally, in the \textit{w. Shuffled Frames} setting, we randomly shuffle the order of the input point cloud frames to evaluate whether \names relies on the correct temporal ordering of motion sequences. In this case, we observe a significant drop of 11.44\% in motion-to-text R@1 under the batch evaluation setting.

Furthermore, to evaluate the robustness of \names under real-world point cloud degradations prevalent in LiDAR sensors, we tested against partially captured point clouds obtained via single-viewpoint back-face culling \cite{partialview} and under random reconstruction errors (normalised std.=0.01) in Table \ref{tab:cl4d_ablations}. \name-mini remains robust under both conditions, losing only a modest 0.37\% and 1.33\% in text-to-motion R@1 under the batch evaluation setting for partial point clouds and random reconstruction errors, respectively.

\begin{figure}[!h]
    \centering
    \begin{minipage}[t]{0.45\linewidth}
        \includegraphics[width=\linewidth]{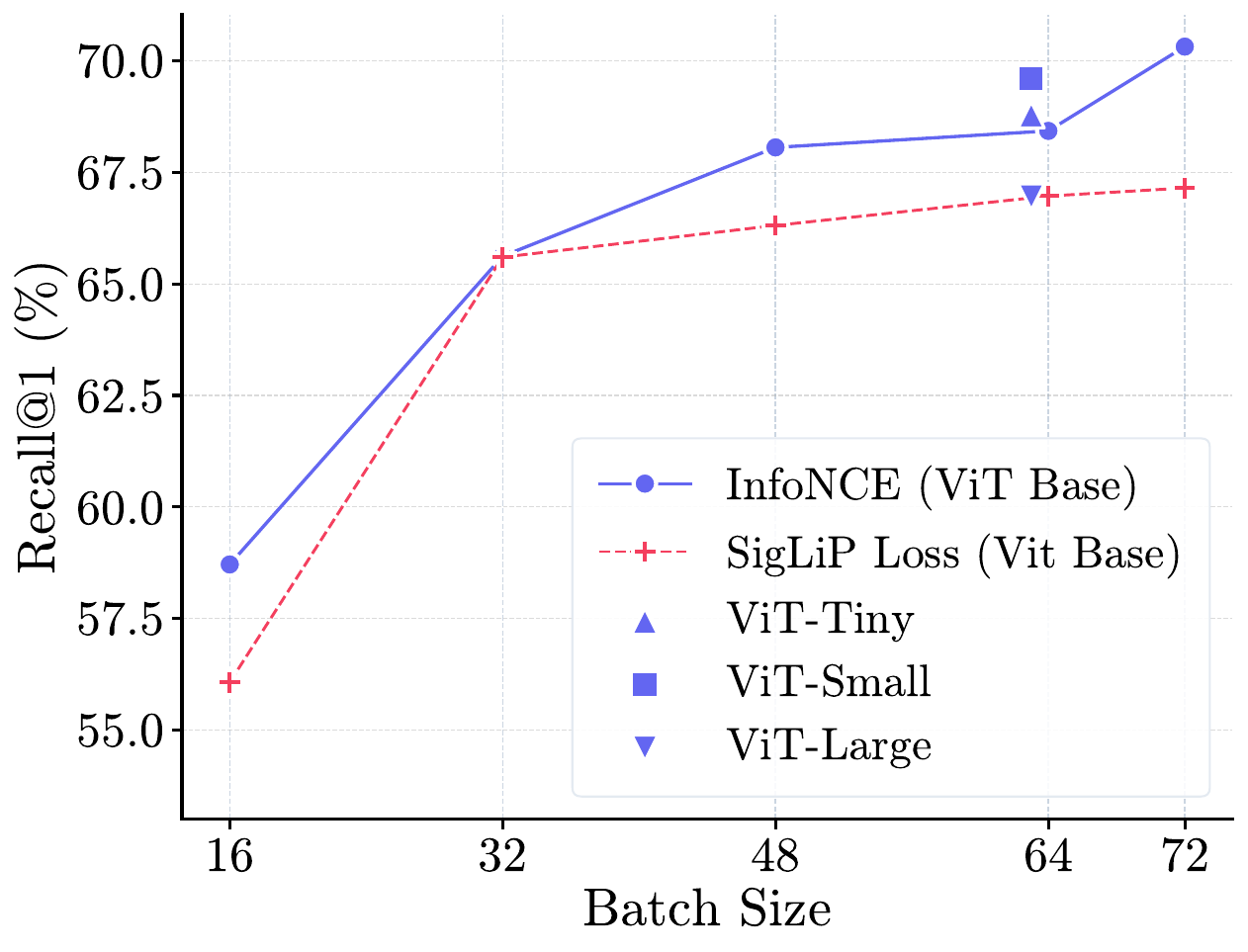}
        \caption{Training batch size vs. Batchwise Recall@1 on text-to-motion retrieval, with results referenced in Table \ref{tab:cl4d_ablations}. Eval. batch size = 32.}
        \label{fig:batchwise_recall_text_to_motion}
    \end{minipage}
    \hfill
    \begin{minipage}[t]{0.45\linewidth}
        \includegraphics[width=\linewidth]{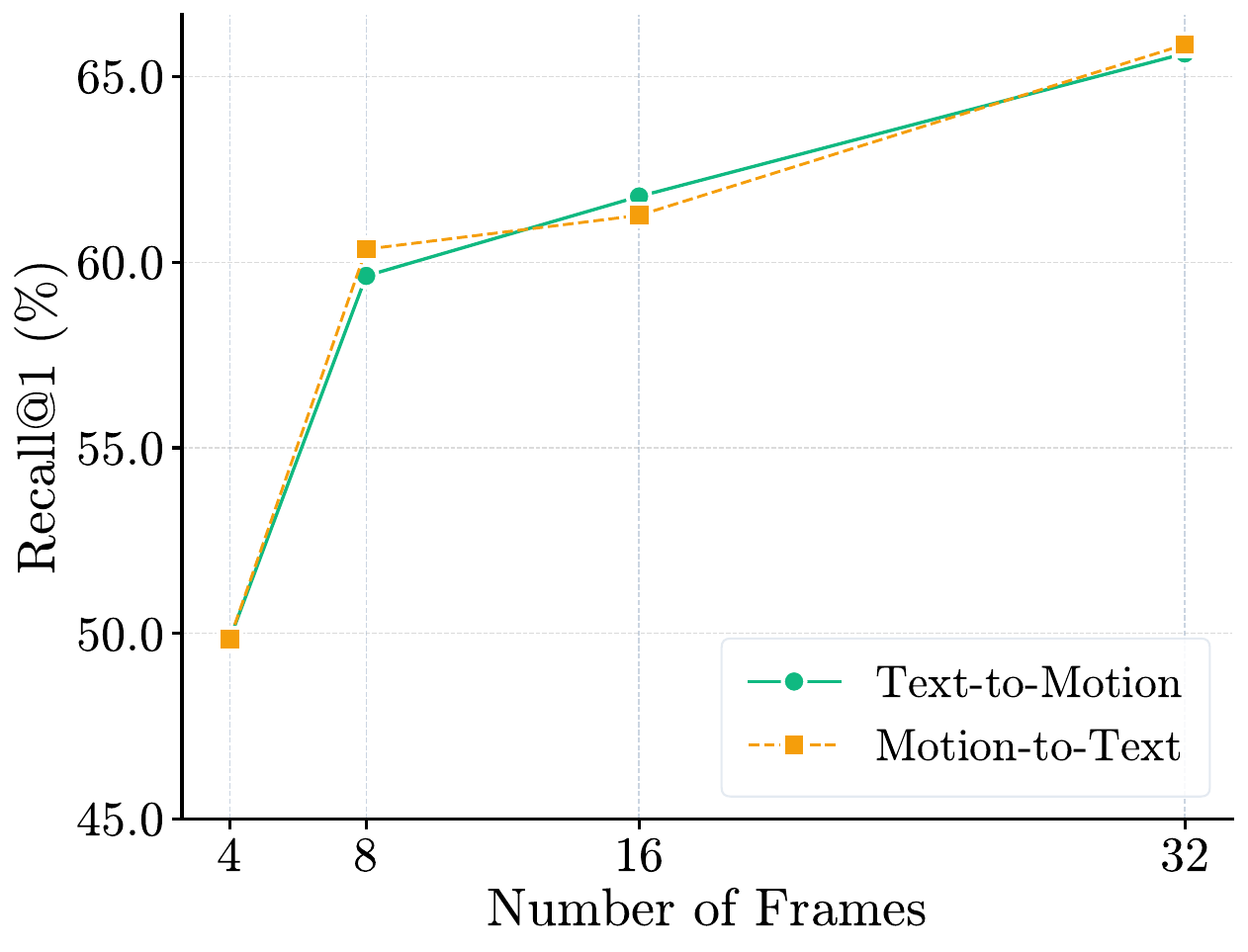}
        \caption{Effect of num. frames for text-to-motion and motion-to-text retrieval evaluated using Batchwise Recall@1. Eval. Batch size = 32.}
        \label{fig:batchwise_num_frames_text_to_scene}
    \end{minipage}
    \vspace{-0.3in}
\end{figure}

In Figure~\ref{fig:batchwise_recall_text_to_motion}, we vary the training batch size used for contrastive pretraining from $B=16$ to $B=72$. Across all configurations of \names, we observe a consistent improvement in R@1 as the batch size increases. This trend can be attributed to the larger number of negative samples available in each batch, which provides stronger contrastive supervision and leads to improved alignment between textual and 4D scene embeddings in the shared representation space.

In Figure~\ref{fig:batchwise_num_frames_text_to_scene}, we vary the number of point cloud frames used as input to \names. For both motion-to-text and text-to-motion retrieval tasks, we observe a consistent increase in R@1 as the number of frames increases. This improvement can be attributed to the richer temporal information captured with a higher number of frames, indicating that \names is able to effectively model temporal dynamics for improved cross-modal retrieval performance.

\begin{table*}[ht]
\centering
\caption{ \textbf{Motion-text Retrieval results on the \dataset-HumanOnly dataset comparing our model (CL4D)  and other experimental setups.} The results show Recall@1 (R@1 $\%$) for both Batch (eval. batch size 32) and Global settings under text-to-motion and motion-to-text retrieval with training batch size = 64. For other variations, we study the effect of freezing the text encoder, using a randomly initialized ViT encoder, shuffling the frames in the point cloud sequence, and optimizing the spatial and temporal encoder, with ViT-Base used as the baseline for all experiments. Under P.Cloud variations, we evaluate \name-mini under real-world point cloud degradations prevalent in LiDAR sensors.}
\label{tab:cl4d_ablations}
\resizebox{\linewidth}{!}{%
\begin{tabular}{llccccc}
\toprule
\multirow{2}{*}{} & \multirow{2}{*}{Model Variation}
  & \multicolumn{2}{c}{Text to Motion}
  & \multicolumn{2}{c}{Motion to Text} & \multirow{2}{*}{\makecell{FLOPs$\downarrow$ \\ ($10^9$)}} \\
\cmidrule(lr){3-4} \cmidrule(lr){5-6}
  & & \makecell{R@1$\uparrow$ \\ (Batch)} & \makecell{R@1$\uparrow$ \\ (Global)} & \makecell{R@1$\uparrow$ \\ (Batch)} & \makecell{R@1$\uparrow$ \\ (Global)} & \\
\midrule
\multirow{5}{*}{\makecell{ViT \\ Variants}}
  & w. ViT-Tiny                     & 68.79 & \textbf{5.15} & 68.09 & 4.24 & 176.20 \\
  & w. ViT-Small                    & \textbf{69.61} & 4.80 & \textbf{70.41} & 4.59 & 177.64 \\
  & w. ViT-Base           & {68.43} & 4.45 & 67.96 & 4.68 & 183.28 \\
  & w. ViT-Large                    & 66.95 & 4.66 & 66.31 & 4.68 & 202.42 \\
\midrule
\multirow{4}{*}{\makecell{Other Model \\ Variations}}
  & w. Frozen Text Enc.          & 56.63 & 2.76 & 57.39 & 2.78 & 183.28 \\
  & w. SigLIP Loss              & 66.97 & 4.03 & 67.14 & 4.20 & 183.28 \\
  & w. untrained ViT      & 63.32 & 3.52 & 62.79 & 3.36 & 183.28 \\
  & w. Shuffled frames               & 60.97 & 3.08 & 58.97 & 2.97 & 183.28 \\
  & CL4D-mini            & 67.18 & 4.24 & 66.02 & \textbf{4.78} & \textbf{70.22} \\

\midrule
\multirow{2}{*}{\makecell{P.Cloud Variations \\ (\name-mini)}}
  & w. Partial P.Cloud         & 66.81 & 4.33 & 65.66 & 4.52 & 70.22 \\
  & w. random recon. error              & 65.85 & 3.69 & 64.64 & 3.82 & 70.22 \\
             
\bottomrule
\end{tabular}%
}
\end{table*}



    

\begin{figure}[!h]
    \centering
    \includegraphics[width=1.0\linewidth]{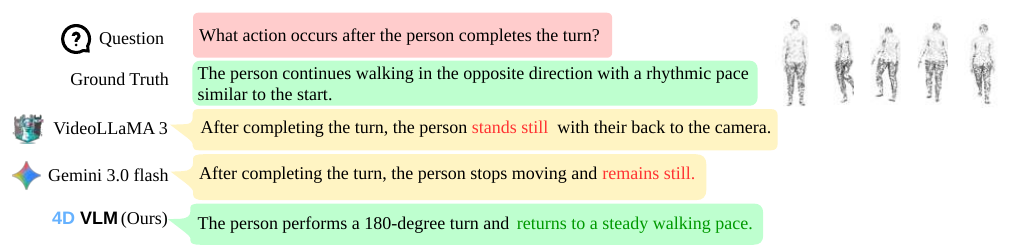}
    \caption{\textbf{Generated outputs} of our 4DVLM on sample VQA instances from \dataset{}-VQA segment, depicting our model's ability to answer questions about human action sequences with significantly higher accuracy and contextual awareness than existing video-based models.}
    \label{fig:qna_diagram}

\end{figure}
\subsection{Evaluation of \llms}

\subsubsection{Evaluation Setup: } We evaluate the performance of \llms with the proposed benchmark dataset \dataset's VQA segment. We primarily use standard VLM benchmark metrics; BLEU \cite{bleu}, ROUGE \cite{lin2004rouge}, METEOR \cite{banerjee-lavie-2005-meteor}, BERTScore \cite{bertscore} and SimCSE \cite{simcse} on VQA with a 4D point cloud as an input.

\subsubsection{Comparison with the State-of-the-art Video VLMs}

\begin{table}[!h]
    \centering
    \caption{\textbf{Visual Question Answering (VQA) results on \dataset{}-VQA, 
comparing our 4DVLM with existing video-based VLMs.} Since 
video models cannot directly process point cloud sequences, 
we render the point clouds as mesh videos and feed them into 
the respective video models for a fair comparison. Our model 
outperforms all video-based baselines across the evaluated metrics.}
\label{tab:evaluation_video}
    \renewcommand{\arraystretch}{1.2}
    \resizebox{\textwidth}{!}{%
    \begin{tabular}{@{}lccccccccc@{}}
        \toprule
        \multirow{2}{*}{\textbf{Method}} & \multirow{2}{*}{\textbf{BLEU}$\uparrow$} & \multicolumn{3}{c}{\textbf{ROUGE}} & \multirow{2}{*}{\textbf{METEOR}$\uparrow$} & \multicolumn{3}{c}{\textbf{BERTScore}} & \multirow{2}{*}{\textbf{SimCSE}$\uparrow$} \\
        \cmidrule(lr){3-5} \cmidrule(lr){7-9}
        & & ROUGE-1(F1)$\uparrow$ & ROUGE-2(F1)$\uparrow$ & ROUGE-L(F1)$\uparrow$ & & Precision$\uparrow$ & Recall$\uparrow$ & F1$\uparrow$ & \\
        \midrule
        VideoLLaMA 3 \cite{zhang2025videollama}&
        $0.0437$ &
        $0.3382$ &
        $0.1174$ &
        $0.2951$ &
        $0.2481$ &
        $\mathbf{0.4769}$ &
        $0.3870$ &
        $0.4317$ &
        $0.8158$ \\
        Gemini-3.0 Flash \cite{geminiteam2025geminifamilyhighlycapable} &
        $0.0447$ &
        $0.3389$ &
        $0.1156$ &
        $0.2812$ &
        $0.2673$ &
        $0.4240$ &
        $0.3875$ &
        $0.4057$ &
        $0.8009$ \\
        Gemini 3.1 Pro \cite{geminiteam2025geminifamilyhighlycapable} &
        $0.0300$ &
        $0.2763$ &
        $0.0923$ &
        $0.2427$ &
        $0.1933$ &
        $0.4562$ &
        $0.2938$ &
        $0.3736$ &
        $0.7690$ \\

        Gemini 3.1 Pro \cite{geminiteam2025geminifamilyhighlycapable}(M.V)  &
        $0.0320$ &
        $0.2885$ &
        $0.0955$ &
        $0.2500$ &
        $0.2064$ &
        $0.4482$ &
        $0.3115$ &
        $0.3788$ &
        $0.7760$ \\
        
        GPT 5 \cite{gpt5} &
        $0.0184$ &
        $0.1935$ &
        $0.0403$ &
        $0.1612$ &
        $0.1347$ &
        $0.4111$ &
        $0.2309$ &
        $0.3191$ &
        $0.7432$ \\

        4DVLM (Ours) &
        $\mathbf{0.0729}$ &
        $\mathbf{0.3857}$ &
        $\mathbf{0.1563}$ &
        $\mathbf{0.3324}$ &
        $\mathbf{0.3152}$ &
        $0.4515$ &
        $\mathbf{0.4396}$ &
        $\mathbf{0.4459}$ &
        $\mathbf{0.8189}$ \\
        \bottomrule
    \end{tabular}
    }
    \vspace{-0.2in}
\end{table}

In Table~\ref{tab:evaluation_video}, we compare \llms against several frontier VLM models. Notably, none of the prior methods are capable of directly consuming 4D point clouds for vision--language reasoning. Therefore, for all baseline VLMs we render the point clouds as mesh videos and feed them into the respective video models for a fair comparison. To provide video baselines with some 3D geometric understanding, we additionally compare against Gemini 3.1 Pro provided with three-viewpoint renders of the same scene, denoted as Gemini 3.1 Pro (M.V) in Table~\ref{tab:evaluation_video}.

Overall, our approach consistently outperforms these frontier VLM baselines across lexical, semantic, and embedding-based evaluation metrics. These results suggest that geometry-aware 4D representations provide stronger grounding for action-centric language generation compared to RGB-based video representations. In Figure~\ref{fig:qna_diagram}, we showcase an example output from \llm, when compared against frontier video-based VLMs.


\subsubsection{Comparison with Prior 4D Encoders: }

\begin{table}[!h]
\vspace{-0.4in}
    \centering
    \caption{\textbf{Visual Question Answering (VQA) results on \dataset{}-VQA, comparing the 4D encoder in our 4DVLM against baseline 4D encoders.} We replace the CL4D encoder in 4DVLM with alternative 4D encoders while 
keeping the projection layer the same across all experimental 
setups. Our method shows consistent gains across all metrics, 
demonstrating the effectiveness of our pretraining approach.
    }
    \label{tab:vlm_encoders}
    \renewcommand{\arraystretch}{1.2}
    \resizebox{\textwidth}{!}{%
    \begin{tabular}{@{}lccccccccc@{}}
        \toprule
        \multirow{2}{*}{\textbf{Method}} & \multirow{2}{*}{\textbf{BLEU}$\uparrow$} & \multicolumn{3}{c}{\textbf{ROUGE}} & \multirow{2}{*}{\textbf{METEOR}$\uparrow$} & \multicolumn{3}{c}{\textbf{BERTScore}} & \multirow{2}{*}{\textbf{SimCSE}$\uparrow$} \\
        \cmidrule(lr){3-5} \cmidrule(lr){7-9}
        & & ROUGE-1(F1)$\uparrow$ & ROUGE-2(F1) $\uparrow$ & ROUGE-L(F1)$\uparrow$ & & Precision$\uparrow$ & Recall$\uparrow$ & F1$\uparrow$ & \\
        \midrule
        w. PST-Transformer &
        $0.0394$ &
        $0.2910$ &
        $0.0938$ &
        $0.2554$ &
        $0.2272$ &
        $0.3808$ &
        $0.3300$ &
        $0.3558$ &
        $0.7551$ \\
        w. Motion PointNet &
        $0.0320$ &
        $0.2932$ &
        $0.0897$ &
        $0.2417$ &
        $0.2513$ &
        $0.3378$ &
        $0.3594$ &
        $0.3488$ &
        $0.7738$ \\
        w. \names (Ours) &
        $\mathbf{0.0729}$ &
        $\mathbf{0.3857}$ &
        $\mathbf{0.1563}$ &
        $\mathbf{0.3324}$ &
        $\mathbf{0.3152}$ &
        $\mathbf{0.4515}$ &
        $\mathbf{0.4396}$ &
        $\mathbf{0.4459}$ &
        $\mathbf{0.8189}$ \\
        \bottomrule
    \end{tabular}
    }
    \label{tab:evaluation_encoders}
    \vspace{-0.2in}
\end{table}

Now, we compare the performance of \llms when using prior 4D encoders to generate 4D scene tokens instead of the proposed \names, as shown in Table~\ref{tab:vlm_encoders}. Although these prior encoders were originally trained using a standard cross-entropy objective, we retrain them with the same contrastive alignment objective for a fair comparison. Even under this setting, \llms with \names consistently achieves higher performance across all evaluation metrics. Among the evaluated 4D encoders, PST-Transformer attains the closest performance to \names. Nevertheless, \llms equipped with \names achieves an 85.03\% higher BLEU score compared to PST-Transformer.

\section{Discussion}

\subsubsection{Rationale for Choosing 4D Pointclouds for Motion Representation: }

Prior work on human motion understanding has primarily relied on skeleton-based representations, which, while effective for human-only motion, cannot generalise to arbitrary objects or non-human agents. In our ObjInteractions segment, there is no well-defined skeletal representation for the 73 interacting object categories, making skeleton-based methods fundamentally inapplicable. Raw 4D point clouds instead provide a unified representation that naturally encodes the geometry and motion of any dynamic entity, as further evidenced by our results on robotic manipulation sequences. With \name, we aim for a general 4D encoder for motion understanding across both human and non-human agents, as demonstrated by the \dataset-ObjInteractions segment and RH20T dataset, with broader generalisation remaining an important direction for future work.

\section{Conclusion}

We present \name, a foundational 4D vision encoder that directly operates on temporally evolving 3D point clouds to generate spatio-temporal vision tokens capturing both geometric structure and motion dynamics. Unlike prior 4D encoders that rely on cross-entropy classification over a fixed set of action labels, \names employs a contrastive learning objective that aligns 4D scene representations with their corresponding language descriptions in a shared embedding space. This design enables semantically grounded 4D representations and significantly improves cross-modal retrieval performance. In particular, \names achieves improvements of 8.1--16.75\% in batch Recall@1 for the text-to-motion retrieval task compared to prior 4D encoders. Building on this encoder, we further introduce \llm, the first vision--language model designed to directly reason over dynamic 4D point clouds. By conditioning language generation on geometry-aware spatio-temporal features, \llms enables language-grounded understanding of dynamic scenes. Experiments demonstrate that \llms achieves superior performance compared to frontier video-based VLMs, including GPT-5 and Gemini, even when these models are provided with corresponding RGB video.

\section{Acknowledgments}


A part of the computational resources were from the funding provided by Manoj Bandara. Also the research was supported by the LK Domain Registry and the Singapore National Research Foundation (NRF), Prime Minister’s Office, Singapore, under its Campus for Research Excellence and Technological Enterprise (CREATE) programme. The Mens, Manus, and Machina (M3S) is an interdisciplinary research group (IRG) of the Singapore-MIT Alliance for Research and Technology (SMART) centre.


\bibliographystyle{splncs04}
\bibliography{main}
\newpage

\section{Appendix}
\subsection{Point Cloud Sequence generation }
\paragraph{SMPL Parameter Conversion:}  

Using the pose parameters from HumanML3D~\cite{humanml3d}, we derive SMPL parameters using PriorMDM~\cite{shafir2024human}. Using these SMPL parameters we generate dynamic human point cloud sequences to construct the 4D representations as explained in the following. 

The original SMPL parameters are provided as:
\begin{itemize}
    \item Joint rotations: $\boldsymbol{\theta} \in \mathbb{R}^{24 \times 6 \times T}$ (rotation-6D representation),
    \item Root translation: $\mathbf{t} \in \mathbb{R}^{3 \times T}$,
    \item Shape parameters: $\boldsymbol{\beta} \in \mathbb{R}^{10}$.
\end{itemize}

Since Unity requires quaternion-based rotations, we convert the $(24,6,T)$ rotation-6D representation into quaternion format $(24,4,T)$. The original motion data follows a right-handed, Z-up coordinate system, whereas Unity uses a left-handed, Y-up coordinate system. Therefore, we apply the following coordinate and rotation transformations to ensure consistency between the two systems:

\begin{itemize}
    \item Translation: $(x,y,z) \rightarrow (-x, z, -y)$
    \item Quaternion: $(x,y,z,w) \rightarrow (-x, y, z, -w)$
\end{itemize}

The 24 SMPL joint names are mapped to indexed bone identifiers and loaded frame-by-frame into the SMPL mesh renderer within Unity.

\paragraph{Scene Placement and Augmentation:}  

Each training animation is randomly placed within a $10 \times 10$ m area and assigned a random global rotation, serving as a form of spatial augmentation. In contrast, test sequences are positioned at the origin $(0,0,0)$ without additional rotation.

For rendering diversity, each animation is randomly assigned either a male or female SMPL mesh, a random shape from a set of predetermined shape parameters   and a texture which is sampled from the SURREAL dataset \cite{varol2017learning}.

\paragraph{Point Cloud Rendering and Surface Sampling:}  

Animations are rendered at 20 FPS. For each frame, surface points are uniformly sampled to obtain approximately 2048 points per human instance. Although the rendering process produces RGB point clouds $(r,g,b,x,y,z)$, our model uses only the geometric coordinates $(x,y,z)$ during training and inference. This removes reliance on appearance cues and supports privacy-preserving deployment.
We adopt area-weighted uniform surface sampling as follows:

\emph{Step 1: Surface Area Computation.}

For each triangle in the mesh, we compute its surface area and store vertex coordinates. The total surface area $A_{\text{total}}$ is obtained by summation. The number of sampled points is computed as:
\[
N = \text{round}(A_{\text{total}} \times \text{density}),
\]
where the density is chosen to yield approximately 2048 points. A cumulative distribution function (CDF) is constructed from per-triangle areas.

\emph{Step 2: Triangle Selection.}  
For each of the $N$ points, we sample $r \sim \mathcal{U}(0, A_{\text{total}})$ and use binary search over the CDF to select a triangle proportional to its area.

\emph{Step 3: Barycentric Sampling.}  
Given triangle vertices $A, B, C$, two random variables $u,v \sim \mathcal{U}(0,1)$ are sampled. If $u+v>1$, we set $u=1-u$ and $v=1-v$. The surface point is then computed as:
\[
P = A + u(B-A) + v(C-A),
\]
producing uniformly distributed points across the mesh surface.
\subsection{DynAction4D-Cluttered Environment Creation}

\subsubsection{Environment Generation and Clutter Synthesis}
To synthesize diverse and realistic cluttered scenes, we curated a collection of 61 common household objects sourced from the Humoto dataset. To ensure scale variation across scenes, these objects were systematically categorized into three groups based on their bounding volumes: \textit{large} ($n=16$), \textit{medium} ($n=21$), and \textit{small} ($n=24$). The specific objects comprising each category are detailed as follows:

\begin{itemize}
    \item \textbf{Large Objects:} bed, dining chair, low chair, working chair, table, side table, kitchen counter, small kitchen counter, utility cart, clothes rack, drawer, shelf, floor lamp, vacuum flask body, trash can, wash tub.
    \item \textbf{Medium Objects:} basket, 90-degree basket, woven basket, medium organizer, small organizer, mango, orange, laptop, laptop top, laptop bottom, mixing bowl, serving bowl, plastic bowl, stacked plastic bowls, wok turner, frying pan, ukulele, guitar, vase, tray, step stool.
    \item \textbf{Small Objects:} spoon, fork, knife, peeler, pen, screwdriver, whisk, turner, soap dispenser, soap dispenser body, soap dispenser top, USB, phone, notebook, mug, deep plate, side plate, cutting board, lint roller, shower squeegee, tap, can, flower, hammer.
\end{itemize}

To rigorously evaluate model generalization to unseen geometries, the total object pool was randomly partitioned into mutually exclusive training (70\%) and testing (30\%) sets.

We procedurally generated 20 distinct training environments and 8 testing environments. To guarantee morphological diversity within each scene, the environment instantiation algorithm mandates the inclusion of at least one randomly sampled object from each of the three size categories (restricted to the respective train/test split). To further increase scene complexity and semantic richness, 1 to 3 additional objects were uniformly sampled from the split's total available object pool and injected into the environment. Duplicate object instances within a single environment were filtered out, yielding unique, procedurally generated clutter compositions for every environment.

\subsubsection{Motion Sequence Distribution}
To construct the Cluttered dataset segment, motion sequences from the HumanML3D dataset were mapped to the synthesized environments. To prevent temporal or stylistic bias during training, the motion sequences corresponding to each split were first randomly shuffled. 

The sequences were then uniformly partitioned across the generated environments. Formally, for a given split containing $S$ total sequences and $E$ generated environments, each environment was assigned approximately $\lfloor S / E \rfloor$ unique sequences. Any residual sequences resulting from indivisibility were allocated to the final environment to ensure exhaustive assignment.

\subsection{Additional Ablation Studies}

\begin{table*}[!h]
\centering
\label{tab:vit_variants}
\caption{\textbf{Motion-text Retrieval results on the \datasets{}- HumanOnly dataset comparing our model (CL4D) and other experimental setups.} The results show Recall@1 (R@1 \%) for both Batch (eval. batch size 32) and Global settings under text-to-motion and motion-to-text retrieval with training batch size = 64. For other variations, we study the effect of freezing the text encoder, using a randomly initialized ViT encoder, shuffling the frames in the point cloud sequence, and optimizing the spatial and temporal encoder, with ViT-Base used as the baseline for all experiments.}
\resizebox{\linewidth}{!}{
\begin{tabular}{lcccccccccccccccccccc}
\toprule
\multirow{3}{*}{} & \multicolumn{10}{c}{Text-to-Motion Retrieval} & \multicolumn{10}{c}{Motion-to-Text Retrieval} \\
\cmidrule(lr){2-11} \cmidrule(lr){12-21}
 & \multicolumn{5}{c}{Global} & \multicolumn{5}{c}{Batchwise} & \multicolumn{5}{c}{Global} & \multicolumn{5}{c}{Batchwise} \\
\cmidrule(lr){2-6} \cmidrule(lr){7-11} \cmidrule(lr){12-16} \cmidrule(lr){17-21}
 & R@1$\uparrow$ & R@2$\uparrow$ & R@3$\uparrow$ & R@5$\uparrow$ & R@10$\uparrow$ & R@1$\uparrow$ & R@2$\uparrow$ & R@3$\uparrow$ & R@5$\uparrow$ & R@10$\uparrow$ & R@1$\uparrow$ & R@2$\uparrow$ & R@3$\uparrow$ & R@5$\uparrow$ & R@10$\uparrow$ & R@1$\uparrow$ & R@2$\uparrow$ & R@3$\uparrow$ & R@5$\uparrow$ & R@10$\uparrow$ \\
\midrule
w. ViT-Tiny & \textbf{5.15} & \textbf{9.25} & \textbf{12.52} & 18.78 & 29.72 & 68.79 & 82.79 & 88.37 & 93.43 & \textbf{96.78} & 4.24 & 8.88 & 12.45 & 18.59 & 28.98 & 68.09 & 83.23 & \textbf{88.84} & 93.36 & \textbf{97.01} \\
 w. ViT-Small & 4.80 & 8.81 & 12.42 & \textbf{19.17} & \textbf{29.95} & \textbf{69.61} & \textbf{83.72} & \textbf{88.87} & \textbf{93.44} & 96.71 & 4.59 & \textbf{9.32} & \textbf{12.68} & \textbf{19.75} & \textbf{31.02} & \textbf{70.41} & \textbf{83.78} & 88.76 & \textbf{93.47} & 96.85 \\
 w. ViT-Base & 4.45 & 9.20 & 12.40 & 17.18 & 28.67 & 68.43 & 81.42 & 86.77 & 91.83 & 96.04 & \textbf{4.68} & 9.11 & 12.47 & 18.20 & 28.58 & 67.96 & 82.10 & 87.62 & 92.03 & 96.16 \\
 w. ViT-Large & 4.66 & 8.99 & 12.15 & 17.83 & 27.86 & 66.95 & 80.81 & 86.53 & 91.42 & 95.90 & \textbf{4.68} & 8.74 & 11.80 & 17.48 & 27.91 & 66.31 & 80.09 & 85.89 & 91.78 & 95.72 \\
\midrule
w. Frozen Text Enc. & 2.76 & 5.19 & 7.05 & 11.17 & 18.73 & 56.63 & 73.17 & 80.19 & 87.72 & 94.81 & 2.78 & 5.15 & 7.53 & 11.50 & 19.84 & 57.39 & 73.71 & 81.19 & 87.79 & 94.93 \\
w. untrained ViT & 3.52 & 6.93 & 9.25 & 13.93 & 22.74 & 63.32 & 78.48 & 84.93 & 90.90 & 96.46 & 3.36 & 6.00 & 9.13 & 13.79 & 23.37 & 62.79 & 78.73 & 84.67 & 90.86 & 96.32 \\
w. Shuffled frames & 3.08 & 5.68 & 8.39 & 12.96 & 21.42 & 60.97 & 76.39 & 82.90 & 89.70 & 96.04 & 2.97 & 5.89 & 8.69 & 12.56 & 20.84 & 58.97 & 74.50 & 82.04 & 89.79 & 96.16 \\
CL4D-mini & 4.24 & 8.58 & 11.85 & 17.80 & 28.12 & 67.18 & 81.14 & 86.82 & 92.04 & 96.64 & 4.78 & 9.23 & 12.12 & 17.52 & 27.65 & 66.02 & 81.68 & 87.58 & 92.78 & 96.94 \\
\bottomrule
\end{tabular}
}
\end{table*}

\begin{table*}[!h]
\centering
\label{tab:loss_n_batchsize_effect}
\caption{\textbf{Motion-text Retrieval results on the \datasets{}-HumanOnly dataset comparing different training batch sizes and different loss functions using Recall@1.}}
\resizebox{\linewidth}{!}{
\begin{tabular}{llcccccccccccccccccccc}
\toprule
\multirow{3}{*}{Loss fn.} & \multirow{3}{*}{Batch size} & \multicolumn{10}{c}{Text-to-Motion Retrieval} & \multicolumn{10}{c}{Motion-to-Text Retrieval} \\
\cmidrule(lr){3-12} \cmidrule(lr){13-22}
 & & \multicolumn{5}{c}{Global} & \multicolumn{5}{c}{Batchwise} & \multicolumn{5}{c}{Global} & \multicolumn{5}{c}{Batchwise} \\
\cmidrule(lr){3-7} \cmidrule(lr){8-12} \cmidrule(lr){13-17} \cmidrule(lr){18-22}
&  & R@1$\uparrow$ & R@2$\uparrow$ & R@3$\uparrow$ & R@5$\uparrow$ & R@10$\uparrow$ & R@1$\uparrow$ & R@2$\uparrow$ & R@3$\uparrow$ & R@5$\uparrow$ & R@10$\uparrow$ & R@1$\uparrow$ & R@2$\uparrow$ & R@3$\uparrow$ & R@5$\uparrow$ & R@10$\uparrow$ & R@1$\uparrow$ & R@2$\uparrow$ & R@3$\uparrow$ & R@5$\uparrow$ & R@10$\uparrow$ \\
\midrule

\multirow{5}{*}{SigLIP-like} & 16 & 3.04 & 5.40 & 7.37 & 11.15 & 18.31 & 56.07 & 72.52 & 79.70 & 87.40 & 94.17 & 2.74 & 5.42 & 7.46 & 11.06 & 18.20 & 55.90 & 71.86 & 78.51 & 86.50 & 93.80 \\
& 32 & 4.20 & 7.79 & 11.06 & 16.90 & 27.93 & 65.60 & 78.73 & 84.47 & 89.73 & 94.07 & 3.92 & 7.83 & 11.27 & 17.41 & 27.82 & 65.41 & 79.01 & 84.43 & 89.43 & 94.51 \\
& 48 & 4.15 & 8.16 & 10.94 & 15.97 & 25.38 & 66.31 & 80.78 & 86.51 & 91.98 & 96.74 & 3.66 & 7.58 & 10.41 & 16.13 & 26.89 & 66.29 & 80.34 & 86.34 & 91.92 & 96.71 \\
& 64 & 4.03 & 7.70 & 10.92 & 17.01 & 27.68 & 66.97 & 81.10 & 87.10 & 92.25 & 96.57 & 4.20 & 8.60 & 11.94 & 17.50 & 28.03 & 67.14 & 81.69 & 86.85 & 91.76 & 96.39 \\
& 72 & 4.15 & 8.14 & 11.85 & 17.85 & 28.93 & 67.15 & 81.19 & 86.44 & 91.71 & 95.90 & 4.29 & 9.04 & 12.22 & 17.78 & 28.91 & 66.66 & 80.22 & 86.41 & 91.73 & 96.32 \\
\midrule
\multirow{5}{*}{\makecell{InfoNCE\\ (Ours)}} & 16 & 2.85 & 5.75 & 7.83 & 12.01 & 19.91 & 58.71 & 73.86 & 81.47 & 87.72 & 94.83 & 2.64 & 5.42 & 7.74 & 11.94 & 19.94 & 58.02 & 73.95 & 80.71 & 87.14 & 94.10 \\
& 32 & 4.82 & 8.90 & 12.15 & 17.55 & 28.81 & 65.62 & 79.00 & 84.35 & 89.78 & 94.42 & 5.08 & 9.06 & 12.24 & 18.27 & 29.79 & 65.86 & 78.39 & 84.12 & 89.85 & 94.65 \\
& 48 & 4.17 & 8.09 & 11.38 & 16.67 & 27.58 & 68.06 & 81.39 & 87.19 & 92.71 & \textbf{96.92} & 3.62 & 8.32 & 11.85 & 17.66 & 27.54 & 67.50 & 81.85 & 88.03 & 92.80 & \textbf{97.08} \\
& 64 & 4.45 & 9.20 & 12.40 & 17.18 & 28.67 & 68.43 & 81.42 & 86.77 & 91.83 & 96.04 & 4.68 & 9.11 & 12.47 & 18.20 & 28.58 & 67.96 & 82.10 & 87.62 & 92.03 & 96.16 \\
& 72 & \textbf{8.07} & \textbf{15.76} & \textbf{21.09} & \textbf{28.98} & \textbf{43.25} & \textbf{70.32} & \textbf{84.10} & \textbf{88.79} & \textbf{92.88} & 96.51 & \textbf{8.02} & \textbf{15.81} & \textbf{20.45} & \textbf{29.30} & \textbf{44.27} & \textbf{68.62} & \textbf{83.27} & \textbf{88.33} & \textbf{92.92} & 96.78 \\
\bottomrule
\end{tabular}
\label{tab:}
}
\end{table*}

\begin{table*}[!h]
\centering
\caption{\textbf{Effect of num. frames for text-to-motion and motion-to-text retrieval evaluated using Batchwise Recall@1.} Eval. Batch size = 32.}
\resizebox{\linewidth}{!}{
\begin{tabular}{ccccccccccccccccccccc}
\toprule
\multirow{3}{*}{Number of Frames} & \multicolumn{10}{c}{Text-to-Motion Retrieval} & \multicolumn{10}{c}{Motion-to-Text Retrieval} \\
\cmidrule(lr){2-11} \cmidrule(lr){12-21}
 & \multicolumn{5}{c}{Global} & \multicolumn{5}{c}{Batchwise} & \multicolumn{5}{c}{Global} & \multicolumn{5}{c}{Batchwise} \\
\cmidrule(lr){2-6} \cmidrule(lr){7-11} \cmidrule(lr){12-16} \cmidrule(lr){17-21}
& R@1$\uparrow$ & R@2$\uparrow$ & R@3$\uparrow$ & R@5$\uparrow$ & R@10$\uparrow$ & R@1$\uparrow$ & R@2$\uparrow$ & R@3$\uparrow$ & R@5$\uparrow$ & R@10$\uparrow$ & R@1$\uparrow$ & R@2$\uparrow$ & R@3$\uparrow$ & R@5$\uparrow$ & R@10$\uparrow$ & R@1$\uparrow$ & R@2$\uparrow$ & R@3$\uparrow$ & R@5$\uparrow$ & R@10$\uparrow$ \\
\midrule
4 & 2.06 & 4.03 & 6.07 & 9.46 & 15.83 & 49.85 & 65.71 & 73.56 & 82.44 & 91.41 & 2.11 & 4.27 & 6.26 & 9.74 & 16.48 & 49.85 & 65.84 & 72.88 & 81.52 & 91.82 \\
8 & 2.94 & 6.19 & 8.83 & 13.17 & 22.35 & 59.63 & 74.84 & 81.09 & 87.76 & 93.95 & 3.59 & 7.32 & 10.20 & 15.35 & 23.60 & 60.35 & 74.68 & 81.45 & 87.25 & 93.83 \\
16 & 3.08 & 6.19 & 8.88 & 13.95 & 23.32 & 61.77 & 75.92 & 82.20 & 88.88 & \textbf{94.46} & 3.43 & 7.26 & 10.27 & 14.58 & 24.08 & 61.26 & 76.29 & 82.22 & 88.30 & 93.95 \\
32 & \textbf{4.82} & \textbf{8.90} & \textbf{12.15} & \textbf{17.55} & \textbf{28.81} & \textbf{65.62} & \textbf{79.00} & \textbf{84.35} & \textbf{89.78} & 94.42 & \textbf{5.08} & \textbf{9.06} & \textbf{12.24} & \textbf{18.27} & \textbf{29.79} & \textbf{65.86} & \textbf{78.39} & \textbf{84.12} & \textbf{89.85} & \textbf{94.65} \\
\bottomrule
\end{tabular}
}
\end{table*}

\begin{table*}[!h]
    \centering
    \caption{\textbf{Motion-text retrieval results on \datasets{}segments.} The results show the Recall@1 for both Batch and Global settings under text-to-motion and motion-to-text retrieval using training batch size = 72 and eval. batch size = 32. CL4D outperforms the other 4D encoders in all the \datasets{}segments.}
    \resizebox{\linewidth}{!}{
    \begin{tabular}{llcccccccccccccccccccc}
    \toprule
    \multirow{3}{*}{\makecell{\datasets \\ Segment}} & \multirow{3}{*}{Method} & \multicolumn{10}{c}{Text-to-Motion Retrieval} & \multicolumn{10}{c}{Motion-to-Text Retrieval} \\
    \cmidrule(lr){3-12} \cmidrule(lr){13-22}
     & & \multicolumn{5}{c}{Global} & \multicolumn{5}{c}{Batchwise} & \multicolumn{5}{c}{Global} & \multicolumn{5}{c}{Batchwise} \\
    \cmidrule(lr){3-7} \cmidrule(lr){8-12} \cmidrule(lr){13-17} \cmidrule(lr){18-22}
    &  & R@1$\uparrow$ & R@2$\uparrow$ & R@3$\uparrow$ & R@5$\uparrow$ & R@10$\uparrow$ & R@1$\uparrow$ & R@2$\uparrow$ & R@3$\uparrow$ & R@5$\uparrow$ & R@10$\uparrow$ & R@1$\uparrow$ & R@2$\uparrow$ & R@3$\uparrow$ & R@5$\uparrow$ & R@10$\uparrow$ & R@1$\uparrow$ & R@2$\uparrow$ & R@3$\uparrow$ & R@5$\uparrow$ & R@10$\uparrow$ \\
    \midrule
    \multirow{4}{*}{HumanOnly} & P4 Transformer & 3.66 & 7.23 & 10.25 & 15.39 & 26.15 & 53.57 & 69.79 & 78.17 & 86.86 & 93.93 & 5.33 & 10.25 & 14.60 & 20.54 & 30.09 & 58.05 & 73.14 & 80.82 & 89.34 & 95.31 \\
    & Pst Transformer & 2.69 & 5.61 & 7.70 & 11.22 & 19.43 & 49.09 & 65.65 & 75.12 & 84.12 & 94.07 & 3.34 & 7.83 & 10.34 & 14.51 & 24.01 & 51.73 & 68.32 & 77.21 & 85.75 & 93.70 \\
    & MotionPointNet & 2.34 & 4.31 & 6.31 & 9.16 & 15.51 & 51.91 & 69.11 & 77.25 & 85.90 & 94.12 & 3.13 & 6.37 & 8.51 & 12.87 & 20.96 & 55.78 & 72.61 & 79.39 & 87.11 & 94.48 \\
    & CL4D (Ours) & \textbf{8.07} & \textbf{15.76} & \textbf{21.09} & \textbf{28.98} & \textbf{43.25} & \textbf{70.32} & \textbf{84.10} & \textbf{88.79} & \textbf{92.88} & \textbf{96.51} & \textbf{8.02} & \textbf{15.81} & \textbf{20.45} & \textbf{29.30} & \textbf{44.27} & \textbf{68.62} & \textbf{83.27} & \textbf{88.33} & \textbf{92.92} & \textbf{96.78} \\
    \midrule
    \multirow{4}{*}{ObjInteractions} & P4 Transformer & 5.94 & 10.05 & 16.44 & 23.74 & 41.10 & 26.55 & 38.49 & 50.35 & 64.88 & 86.00 & 7.31 & 10.50 & 15.98 & 23.29 & 42.47 & 24.93 & 41.78 & 56.32 & 73.69 & 88.59 \\
    & Pst Transformer & 8.22 & 14.61 & 20.09 & 28.31 & 44.75 & 30.37 & 47.39 & 56.04 & 69.68 & 88.59 & 7.31 & 12.79 & 18.72 & 30.59 & 47.49 & 25.50 & 47.06 & 63.62 & 75.84 & 91.72 \\
    & MotionPointNet & 12.33 & 24.20 & 31.96 & 44.29 & 61.19 & 41.37 & 55.54 & 65.08 & 78.97 & \textbf{92.69} & 11.87 & 21.46 & 31.05 & 43.38 & 61.19 & 36.18 & 55.62 & 67.03 & 78.80 & \textbf{93.58} \\
    & CL4D (Ours) & \textbf{23.29} & \textbf{39.27} & \textbf{47.03} & \textbf{56.62} & \textbf{68.04} & \textbf{49.40} & \textbf{63.29} & \textbf{73.00} & \textbf{83.52} & 91.35 & \textbf{22.37} & \textbf{33.33} & \textbf{42.47} & \textbf{53.42} & \textbf{69.86} & \textbf{46.97} & \textbf{63.46} & \textbf{71.58} & \textbf{81.56} & 91.27 \\
    
     \midrule
     \multirow{4}{*}{Cluttered}
    & Pst Transformer & 0.93 & 1.78 & 2.41 & 3.78 & 6.35 & 31.47 & 46.36 & 56.82 & 69.92 & 86.56 & 1.02 & 1.85 & 2.69 & 4.94 & 9.06 & 34.45 & 49.12 & 57.60 & 69.19 & 84.32 \\
    & MotionPointNet & 1.23 & 2.41 & 3.52 & 5.77 & 9.81 & 41.71 & 57.93 & 67.27 & 78.75 & 90.93 & 1.53 & 2.97 & 4.06 & 6.28 & 11.68 & 43.24 & 59.65 & 69.07 & 78.87 & 91.18 \\
    & CL4D (Ours) & \textbf{3.11} & \textbf{6.00} & \textbf{8.48} & \textbf{12.49} & \textbf{20.63} & \textbf{55.07} & \textbf{69.94} & \textbf{77.53} & \textbf{84.58} & \textbf{92.87} & \textbf{2.71} & \textbf{5.54} & \textbf{7.58} & \textbf{11.82} & \textbf{18.94} & \textbf{51.94} & \textbf{67.43} & \textbf{75.77} & \textbf{83.34} & \textbf{91.31} \\

    \bottomrule
    \end{tabular}
    }
    \end{table*}
    
\begin{table}[!h]
    \centering
    \caption{Model Parameters and Computational Complexity for different ViT Variants; Here we can observe the CL4D-mini has the lowest parameter count and computational cost.}
    \begin{tabular}{lrc}
    \toprule
    Model & Training Params $\downarrow$ & FLOPS (Gflops) $\downarrow$ \\ \midrule
    w. ViT-Tiny & 34.65 M & 176.2  \\ 
    w. ViT-Small & 50.82 M & 177.64 \\
    w. ViT-Base  & 115.0 M & 183.28 \\ 
    w. ViT-Large & 332.53 M & 202.42 \\ 
    CL4D-mini                       & \textbf{8.93 M}  & \textbf{70.22}  \\ \bottomrule
    \end{tabular}
    \label{tab:model_comparison_params}
    \end{table}

\begin{figure}[!h]
    \centering
    \begin{minipage}{0.45\linewidth}
        \centering
        \includegraphics[width=\linewidth]{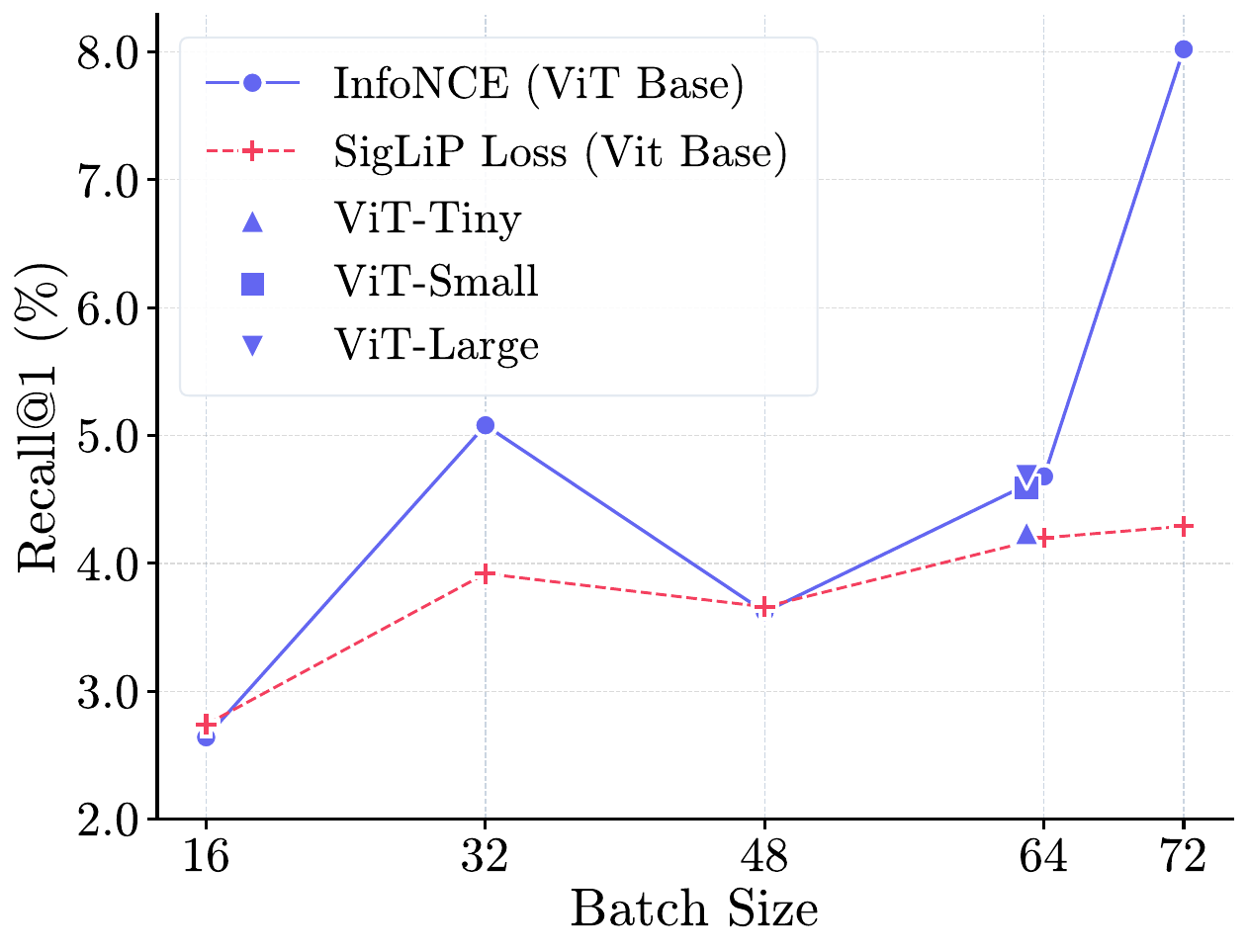}
        \caption{Training batch size vs. Global Recall@1 on motion-to-text retrieval. Eval. batch size = 32.}
        \label{fig:global_motion_to_text}
    \end{minipage}
    \hfill
    \begin{minipage}{0.45\linewidth}
        \centering
        \includegraphics[width=\linewidth]{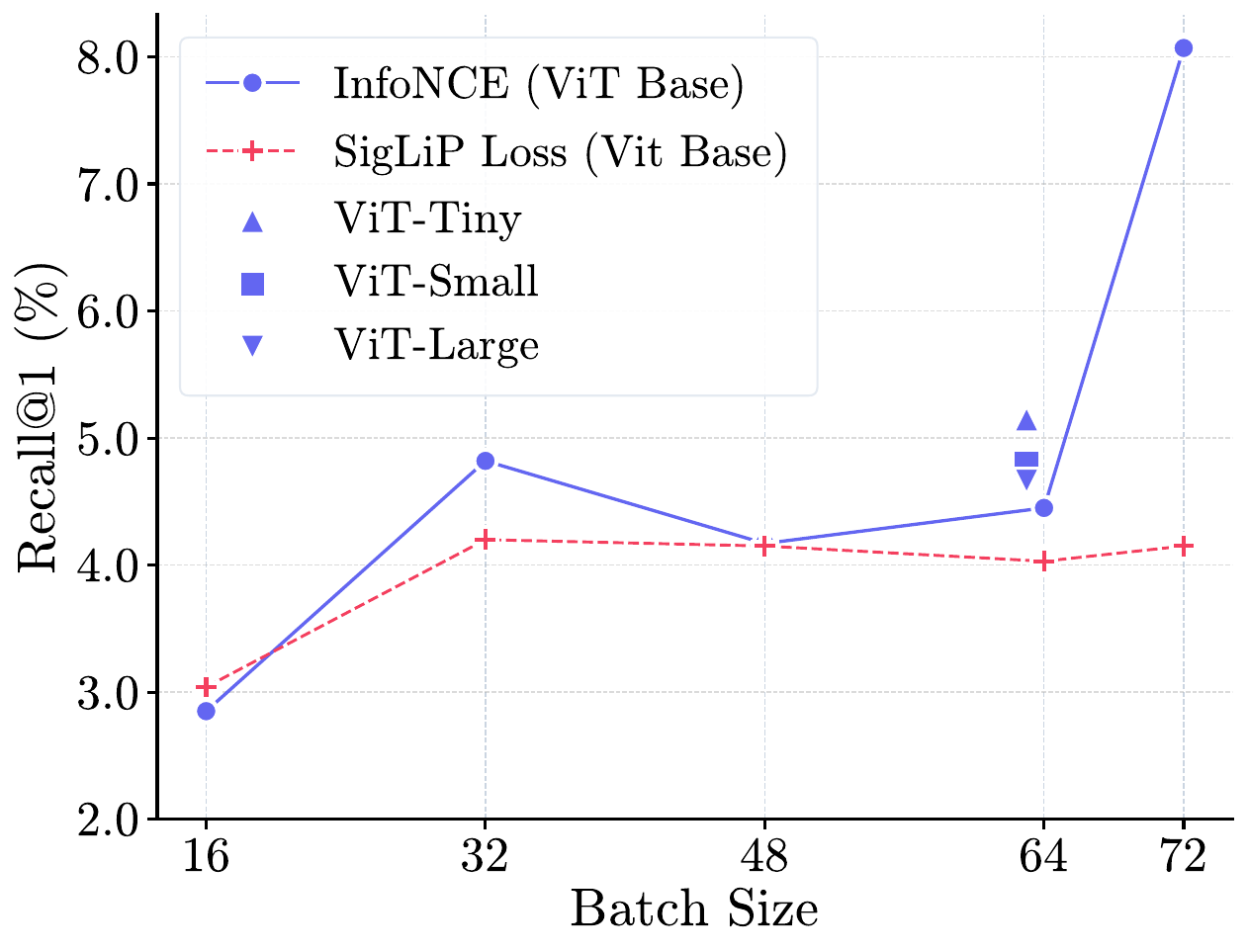}
        \caption{Training batch size vs. Global Recall@1 on text-to-motion retrieval. Eval. batch size = 32.}
        \label{fig:global_text_to_motion}
    \end{minipage}
    \vspace{0.0in}
\end{figure}

\begin{figure}[!h]
    \centering
    \includegraphics[width=0.45\linewidth]{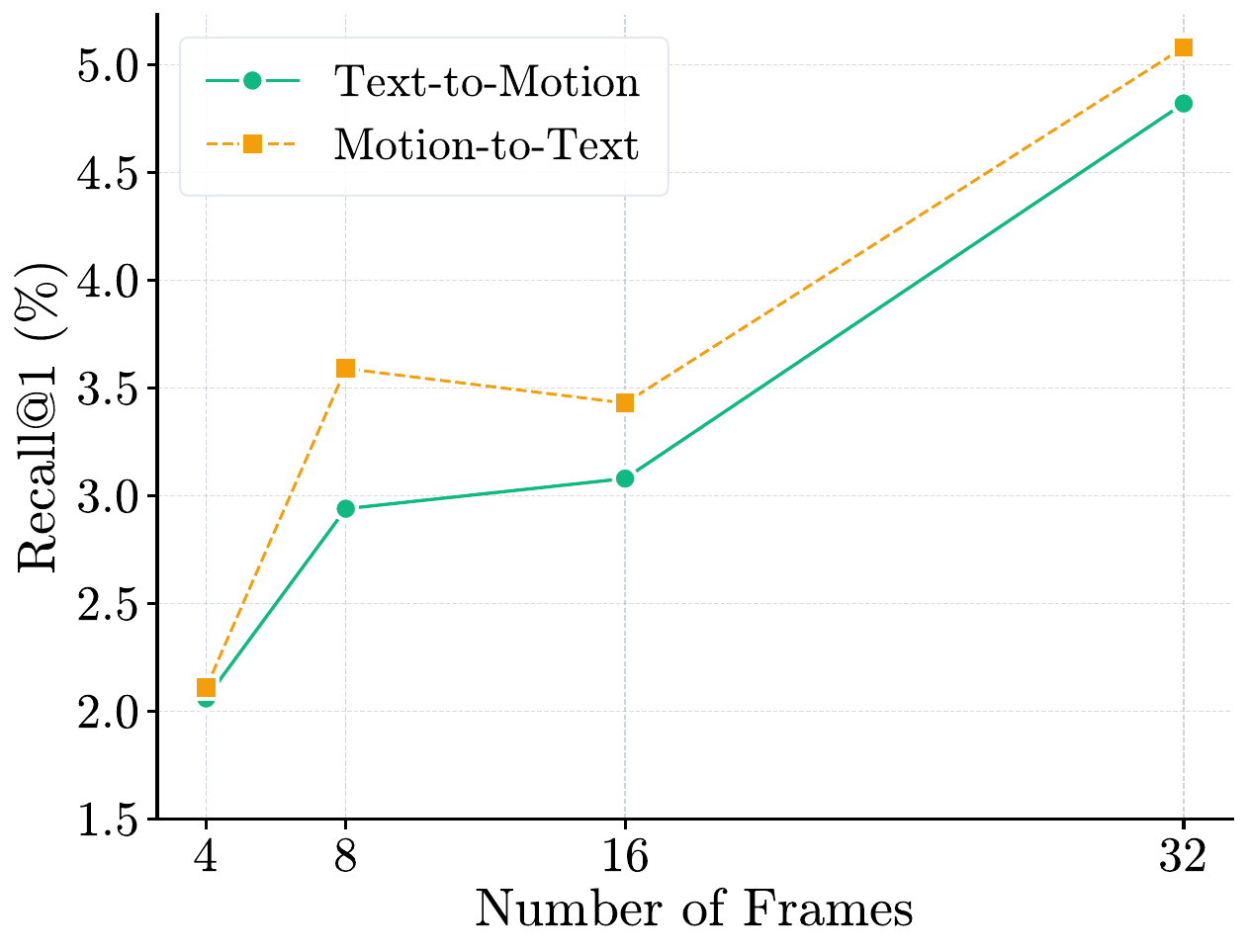}
    \caption{Effect of Num. of frames used in the point cloud sequences for Global Recall@1 text-motion retrieval tasks.}
    \label{fig:global}
    \vspace{0.0in}
\end{figure}

\subsection{Data Generation Prompt}

We first use the following prompt to generate the detailed caption using both the video of the mesh of the point cloud sequence and the HumanML3D dataset caption. We do not consider texture and color information to generate question pairs. Listing~\ref{lst:video_caption_stage1} to Listing~\ref{lst:point_evaluation_prompt} shows the different prompt templates used for data generation and evaluations.

\newpage
\begin{lstlisting}[
  breaklines=true, 
  basicstyle=\ttfamily\scriptsize, 
  frame=single, 
  caption={Prompt template for generating \textbf{detailed captions}}, 
  label={lst:video_caption_stage1},
  ]
### CONTEXT
I am providing {n} consecutive frames from a video derived from the HumanML dataset.
Reference Ground Truth: "{humanML_groundtruth_caption}"

### TASK
Generate a single, detailed English caption that objectively narrates
the movement shown in the frames.

### GUIDELINES
1. **Visual Priority:** Use the Reference Ground Truth ONLY as a high-level
   guide for the context of the action. You must verify that the action
   actually occurs in the specific frames provided. If the visual contradicts
   the text, trust the text.
2. **Subject Description:** Refer to the subject simply as "the person". Do
   not describe their body type, age, face, or clothing unless it is
   mechanically relevant to the interaction (e.g., "holding a skirt").
3. **Motion Focus:** Focus 90 percent of the caption on the mechanics of the
   movement (limbs, posture, speed, trajectory).
4. **Chronology:** Describe the sequence of movements in strict chronological
   order.
5. **Constraints:**
   - NO background description.
   - NO symbolic interpretation (e.g., do not say "he looks sad," say
     "he walks with a slumped posture").
   - NO distinct physical features (hair, eyes, skin).

### OUTPUT
Provide only the caption text.
\end{lstlisting}

\begin{lstlisting}[
  breaklines=true, 
  basicstyle=\ttfamily\scriptsize, 
  frame=single, 
  caption={Prompt template for generating \textbf{QA pairs} from detailed captions}, 
  label={lst:video_caption_stage2}
]
You are given a detailed caption describing a motion sequence from a dataset.
Your task is to generate generalised Question-Answer (QA) pairs derived
*strictly* from the information in the caption.

### CORE PRINCIPLES
1. **One action per question.** Each question must ask about exactly ONE
   action or movement, not a chain of actions.
2. **Keep questions general.** Ask broad questions about what the person
   does, not hyper-specific questions about minute details.
3. **No answer leakage.** The question must NOT contain or hint at the
   answer. The reader should not be able to guess the answer from the
   question alone.

### CATEGORIES
1. **Action:** Ask what the person does during a particular phase.
   - GOOD: "What does the person do at the start of the sequence?"
   - GOOD: "How does the person move after standing up?"
   - BAD: "Does the person raise their left arm?" (answer is leaked)
   - BAD: "How does the person bend their knees while simultaneously
     rotating their torso?" (too specific, multiple actions)
2. **Sequence:** Ask about the order of events without revealing what happens.
   - GOOD: "What does the person do after the first action?"
   - GOOD: "What is the final movement in the sequence?"
   - BAD: "Does the person jump before walking?" (answer leaked)
3. **Body Position:** Ask about body posture or positioning during a movement.
   - GOOD: "What is the position of the person's arms during the movement?"
   - GOOD: "How is the person's body oriented at the end?"
   - BAD: "Are the person's arms raised above the head?" (answer leaked)

### RULES
- Generate 3 to 6 QA pairs per caption.
- If the caption does not contain information for a category, omit that
  category. Do NOT hallucinate.
- The answer must stand alone without needing the question for context.
  Refer to "the person" instead of pronouns.
- Do not copy-paste the caption. Rephrase naturally.
- Avoid yes/no questions. Prefer open-ended "What" / "How" questions.

### INPUT CAPTION
'{generated_caption}'

### OUTPUT FORMAT
Provide the output in valid JSON format only. Do not add markdown backticks.
{
  'qa_pairs': [
    {
      'type': 'Action',
      'question': '...',
      'answer': '...'
    },
    {
      'type': 'Sequence',
      'question': '...',
      'answer': '...'
    },
    {
      'type': 'Body Position',
      'question': '...',
      'answer': '...'
    }
  ]
}
\end{lstlisting}

\begin{lstlisting}[
  breaklines=true, 
  basicstyle=\ttfamily\scriptsize, 
  frame=single, 
  caption={Prompt Template for video evaluation.}, 
  label={lst:video_evaluation_prompt}
]
You are a helpful assistant that analyzes videos. Answer the user's questions based on the video content directly and concisely.
\end{lstlisting}

\begin{lstlisting}[
  breaklines=true, 
  basicstyle=\ttfamily\scriptsize, 
  frame=single, 
  caption={ Prompt Template for point cloud sequence evaluation}, 
  label={lst:point_evaluation_prompt}
]
Default template of Vicuna_V1
\end{lstlisting}


\subsection{Use of Pre-Trained ViT weights for Temporal Encoder initialization}

The use of a pre-trained Vision Transformer (ViT) \cite{vit} weight initialization for our temporal encoder is motivated by its capacity to leverage high-level ``visual knowledge'' and global receptive fields acquired from large-scale image datasets like ImageNet \cite{deng2009imagenet}. In domains like 3D human motion, where high quality annotated data is significantly scarcer than natural images, transfer learning can be used to overcome the data scale problem \cite{mopa}. By stacking spatial features into a grid-like representation, we effectively transform temporal dynamics into a ``motion spectrogram'' or motion image, where complex physical actions transformed as visual textures. This approach follows established methodologies in audio tagging and motion retrieval \cite{psla},\cite{mopa}, which treat time-series sequences as 2D visual inputs to exploit the feature extraction capabilities of pre-trained image models. Unlike traditional recurrent architectures, the ViT’s self-attention mechanism captures global spatial-temporal dependencies and long-range relationships in a single pass.

\subsection{Data Statistics}

\subsubsection{DynAction Dataset Statistics}

\begin{figure}[!h]
    \centering
    \includegraphics[width=0.95\linewidth]{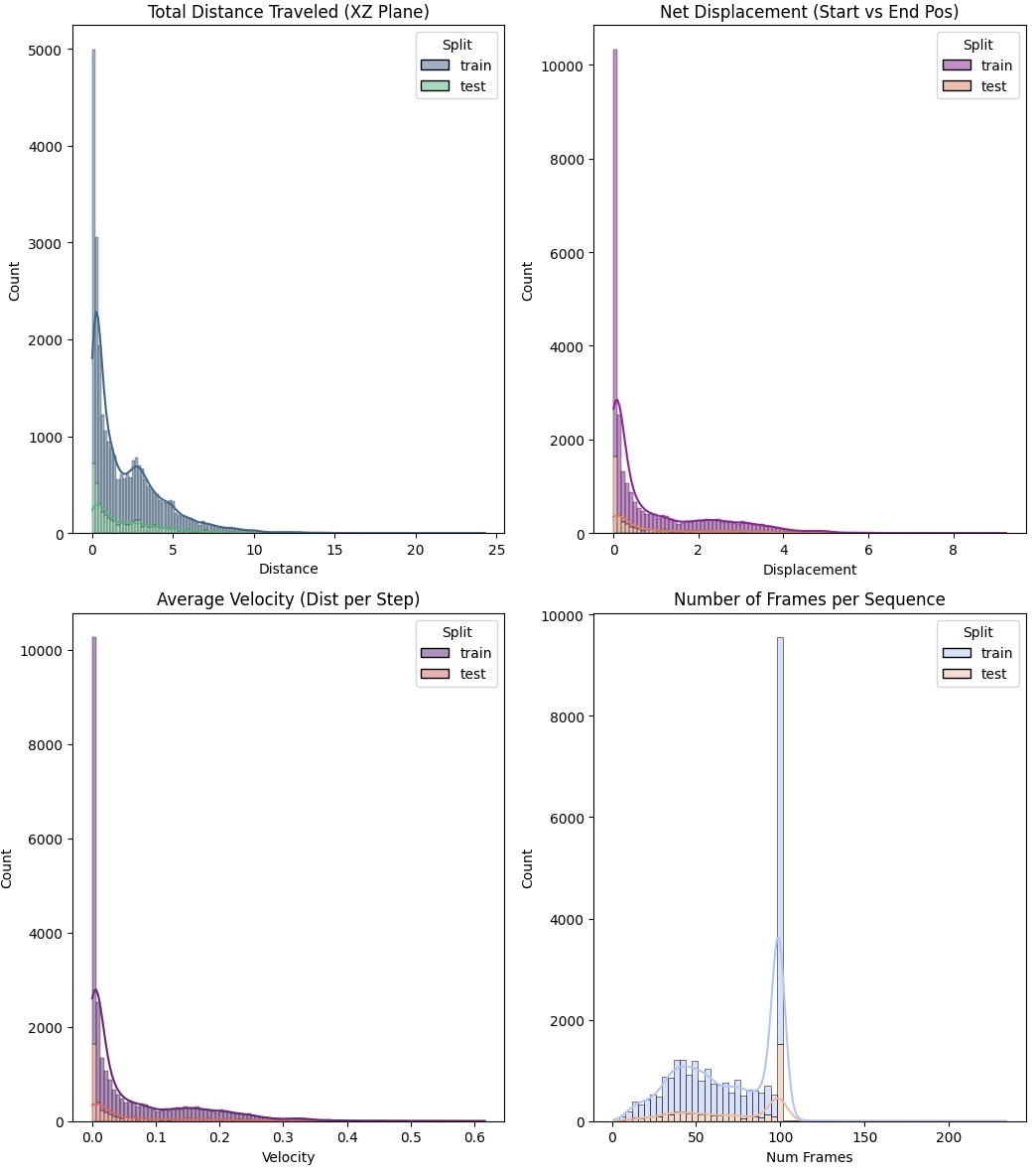}
    \caption{Statistics of the \dataset{}-HumanOnly dataset segment.}
    \label{fig:dynaction_human}
    \vspace{0.0in}
\end{figure}

\begin{figure}[!h]
    \centering
    \includegraphics[width=0.95\linewidth]{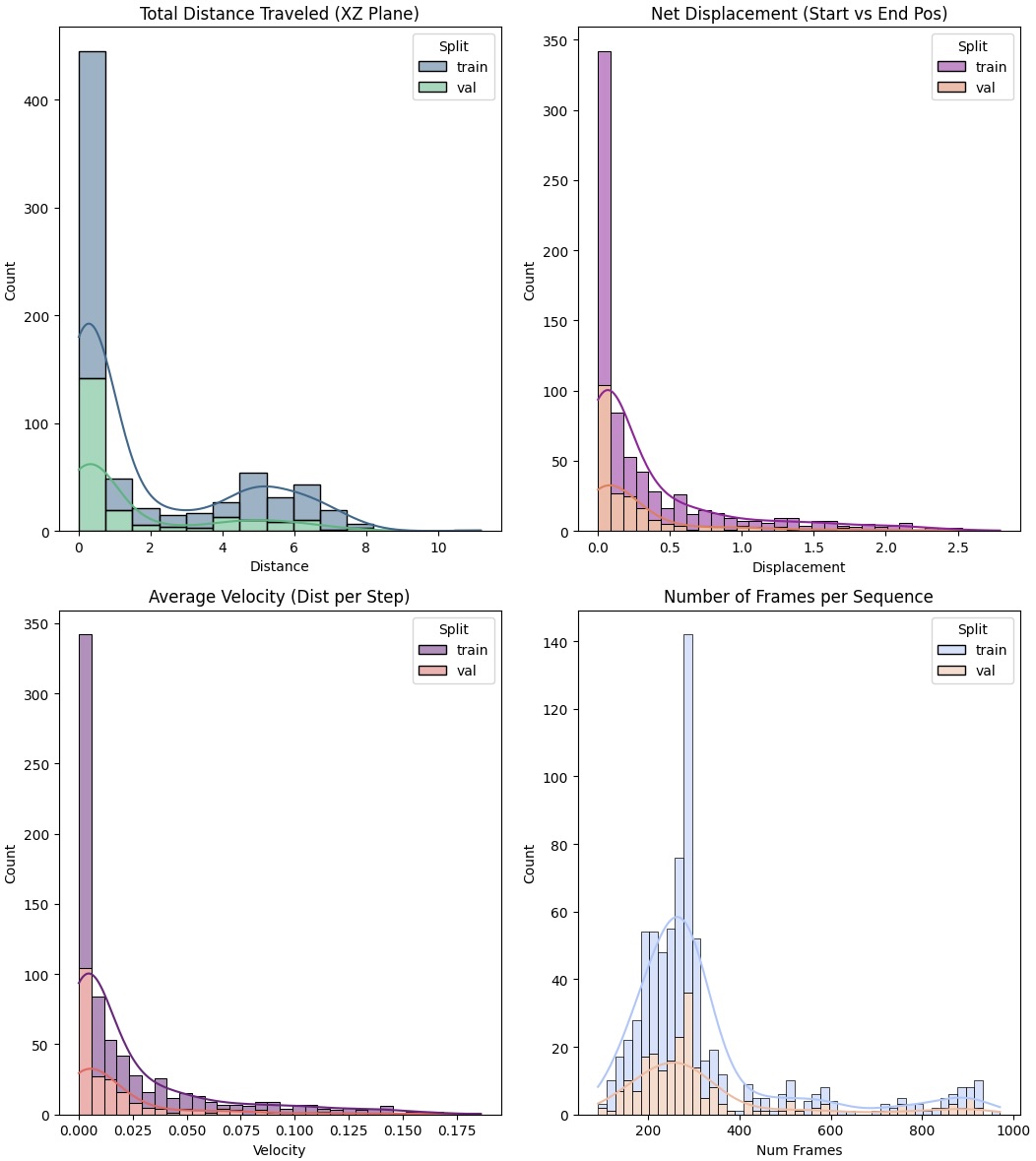}
    \caption{Statistics of the \dataset{}-ObjInteractions dataset segment.}
    \label{fig:dynaction_obj_interactions}
    \vspace{0.0in}
\end{figure}

\begin{figure}[!h]
    \centering
    \includegraphics[width=0.95\linewidth]{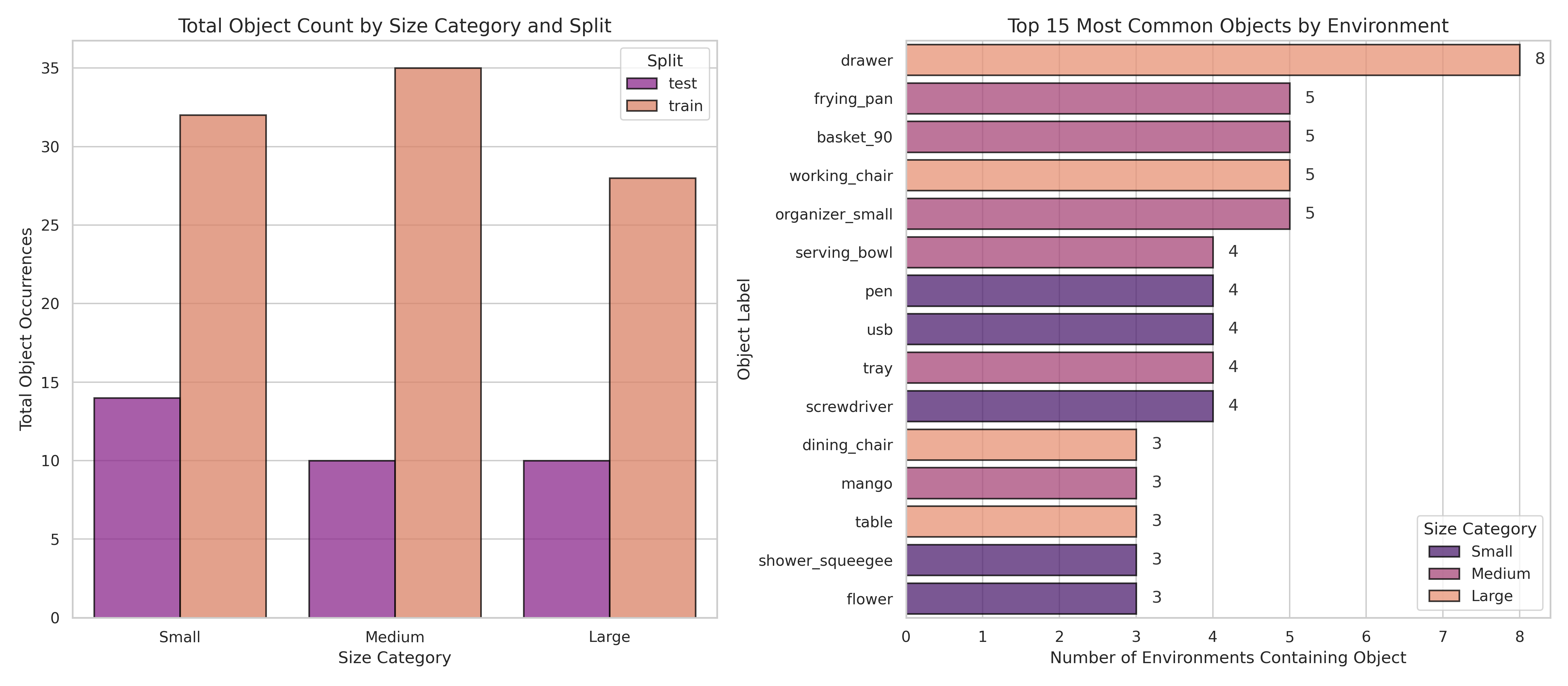}
    \caption{Statistics of the \dataset{}-Cluttered dataset segment.}
    \label{fig:dynaction_combined_obj_interactions}
    \vspace{0.0in}
\end{figure}

The composition of our DynAction dataset segments of \dataset{}-HumanOnly, \dataset{}-ObjInteractions and \dataset{}-Cluttered were shown in the plots Fig.~\ref{fig:dynaction_human} to Fig.~\ref{fig:dynaction_combined_obj_interactions} respectively. Here we analyzed the compositions in terms of human actors' total distance traveled, Net displacement, average velocity, number of frames per each sequence and the composition of the objects.

\subsubsection{DynAction-VQA Dataset training Split Statistics}

The training split contains \textbf{22,965} QA files with a total of \textbf{134,070} question--answer pairs. 
On average, each sequence contains \textbf{5.84} QA pairs (min = 3, max = 9). 
The average number of QA pairs per category per sequence is:
\textit{Action}: 1.94, \textit{Temporal}: 1.82, and \textit{Body-Spatial}: 2.08.

\begin{table}[!h]
\centering
\caption{Word count statistics for questions ($q\_len$) and answers ($a\_len$) in the DynAction-VQA training split.}
\begin{tabular}{lcccccc}
\hline
\textbf{Type} & \multicolumn{3}{c}{$q\_len$} & \multicolumn{3}{c}{$a\_len$} \\
 & Min & Max & Mean & Min & Max & Mean \\
\hline
Action & 5 & 36 & 11.66 & 7 & 55 & 21.66 \\
Body-Spatial & 5 & 23 & 12.82 & 7 & 43 & 18.67 \\
Temporal & 5 & 23 & 12.43 & 7 & 45 & 19.76 \\
\hline
\end{tabular}
\end{table}

\subsubsection{DynAction-VQA Dataset test Split Statistics}

The test split contains \textbf{4,314} QA files with a total of \textbf{25,117} QA pairs. 
Each sequence contains on average \textbf{5.82} QA pairs (min = 3, max = 8). 
The average number of QA pairs per category per sequence is:
\textit{Action}: 1.94, \textit{Temporal}: 1.81, and \textit{Body-Spatial}: 2.06.

\begin{table}[!h]
\centering
\caption{Word count statistics for questions ($q\_len$) and answers ($a\_len$) in the DynAction-VQA test split.}
\begin{tabular}{lcccccc}
\hline
\textbf{Type} & \multicolumn{3}{c}{$q\_len$} & \multicolumn{3}{c}{$a\_len$} \\
 & Min & Max & Mean & Min & Max & Mean \\
\hline
Action & 5 & 26 & 11.68 & 7 & 52 & 21.75 \\
Body-Spatial & 6 & 23 & 12.82 & 7 & 54 & 18.68 \\
Temporal & 5 & 23 & 12.45 & 7 & 41 & 19.76 \\
\hline
\end{tabular}
\end{table}

\subsubsection{Overall DynAction-VQA Dataset Statistics}

Across the full dataset, DynAction-VQA contains \textbf{27,279} QA files and \textbf{159,187} QA pairs. 
Each sequence contains on average \textbf{5.84} QA pairs (min = 3, max = 9). 
The average QA distribution per sequence is:
\textit{Action}: 1.94, \textit{Temporal}: 1.82, and \textit{Body-Spatial}: 2.07.

\begin{table}[!h]
\centering
\caption{Word count statistics for questions ($q\_len$) and answers ($a\_len$) in combined DynAction-VQA dataset.}
\begin{tabular}{lcccccc}
\hline
\textbf{Type} & \multicolumn{3}{c}{$q\_len$} & \multicolumn{3}{c}{$a\_len$} \\
 & Min & Max & Mean & Min & Max & Mean \\
\hline
Action & 5 & 36 & 11.67 & 7 & 55 & 21.67 \\
Body-Spatial & 5 & 23 & 12.82 & 7 & 54 & 18.68 \\
Temporal & 5 & 23 & 12.43 & 7 & 45 & 19.76 \\
\hline
\end{tabular}
\end{table}

\newpage

\subsection{Samples from the \dataset{}-VQA segment and the 4DVLM Predictions}

\begin{figure}[!h] 
    \centering
    
    \begin{minipage}{\linewidth}
        \centering
        \includegraphics[width=\linewidth]{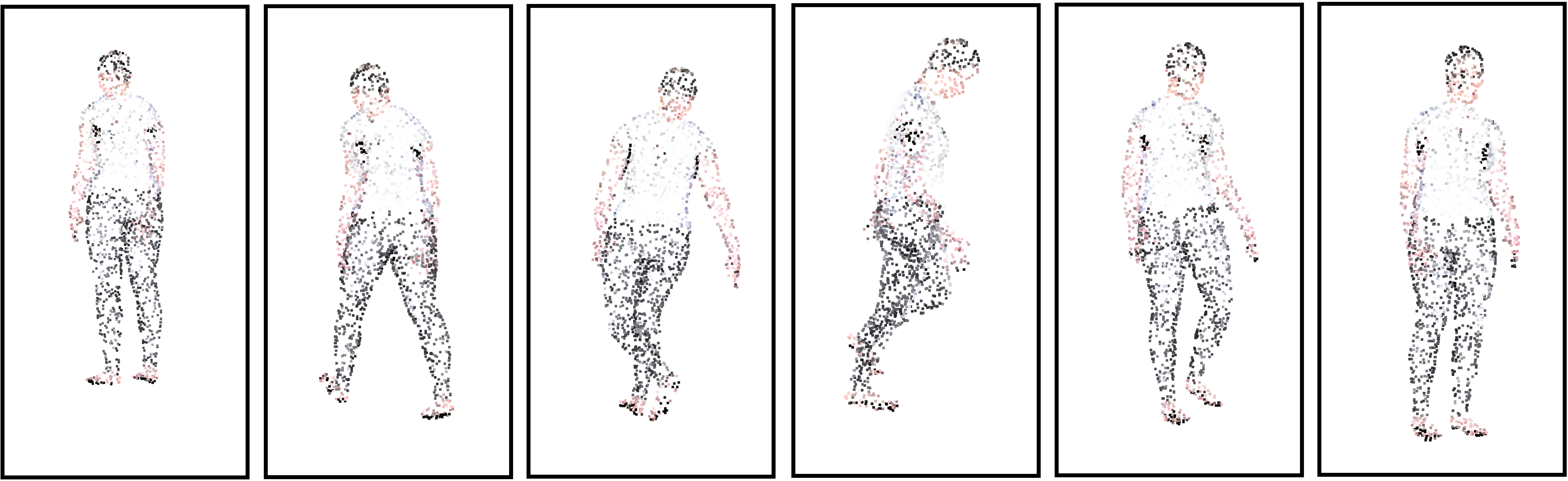}
    \end{minipage}
    
    \vspace{0.3cm} 
    
    \begin{minipage}{\linewidth}
        \centering
        \includegraphics[width=\linewidth]{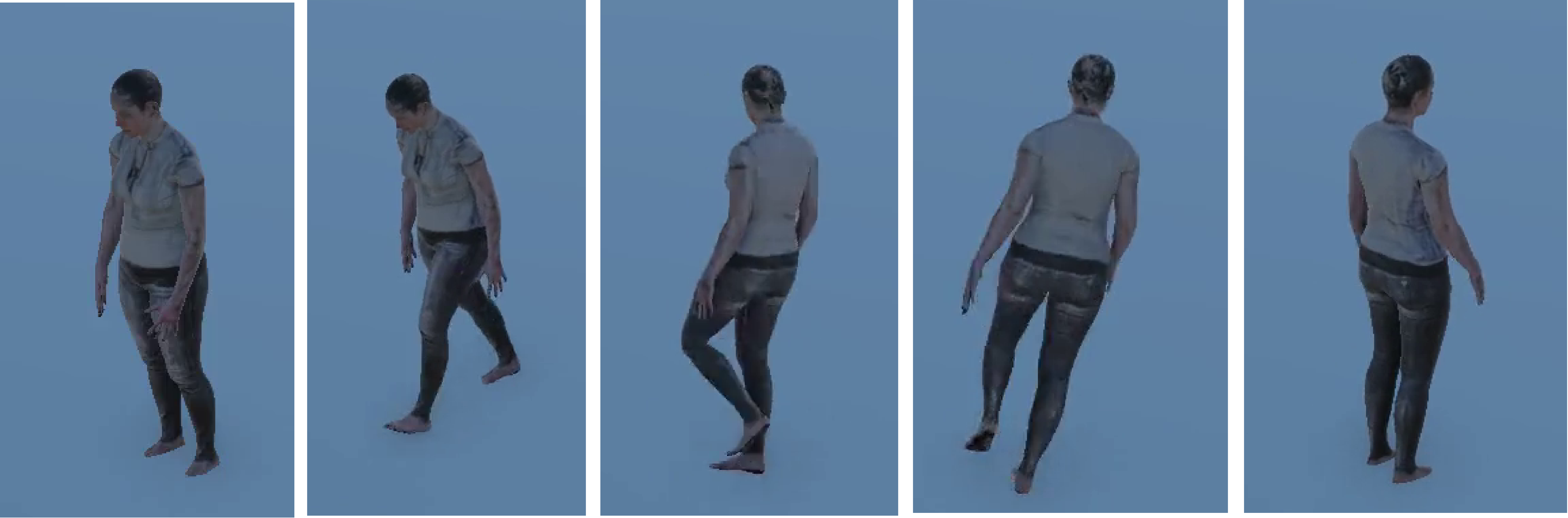}
    \end{minipage}
    
    \vspace{0.4cm} 
    
    \begin{minipage}{\linewidth}
        \small 
        \begin{itemize}
            \setlength{\itemsep}{4pt} 
            
            \item \textbf{Q:} What does the person do at the very beginning of the motion sequence?\\
                  \textbf{GT:} The person walks forward with a balanced posture and coordinated arm movements.\\
                  \textbf{4DVLM:} The person starts from a standing position and begins to walk forward.
                  
            \item \textbf{Q:} What action occurs after the person completes the turn?\\
                  \textbf{GT:} The person continues walking in the opposite direction with a rhythmic pace similar to the start.\\
                  \textbf{4DVLM:} The person performs a 180-degree turn to the right and returns to a steady walking pace.
                  
            \item \textbf{Q:} How does the entire sequence conclude?\\
                  \textbf{GT:} The person gradually decreases speed until coming to a complete stop with both feet aligned together.\\
                  \textbf{4DVLM:} The person returns to a neutral standing position and remains still for a few moments.
                  
            \item \textbf{Q:} How does the person change direction during the sequence?\\
                  \textbf{GT:} The person slows down and performs a 180-degree clockwise turn by pivoting on the balls of the feet.\\
                  \textbf{4DVLM:} The person performs a 180-degree turn to the right while maintaining a steady pace.
        \end{itemize}
    \end{minipage}
    
    \caption{Qualitative results for \texttt{sequence\_000022}. The top row visualizes the point cloud sequence, while the bottom row displays the corresponding rendered video. The corresponding question-answering pairs, ground truth (GT), and model predictions (4DVLM) are provided below.}
    \label{fig:seq_000022}
\end{figure}

\begin{figure}[!h]
    \centering
    
    \begin{minipage}{\linewidth}
        \centering
        \includegraphics[width=\linewidth]{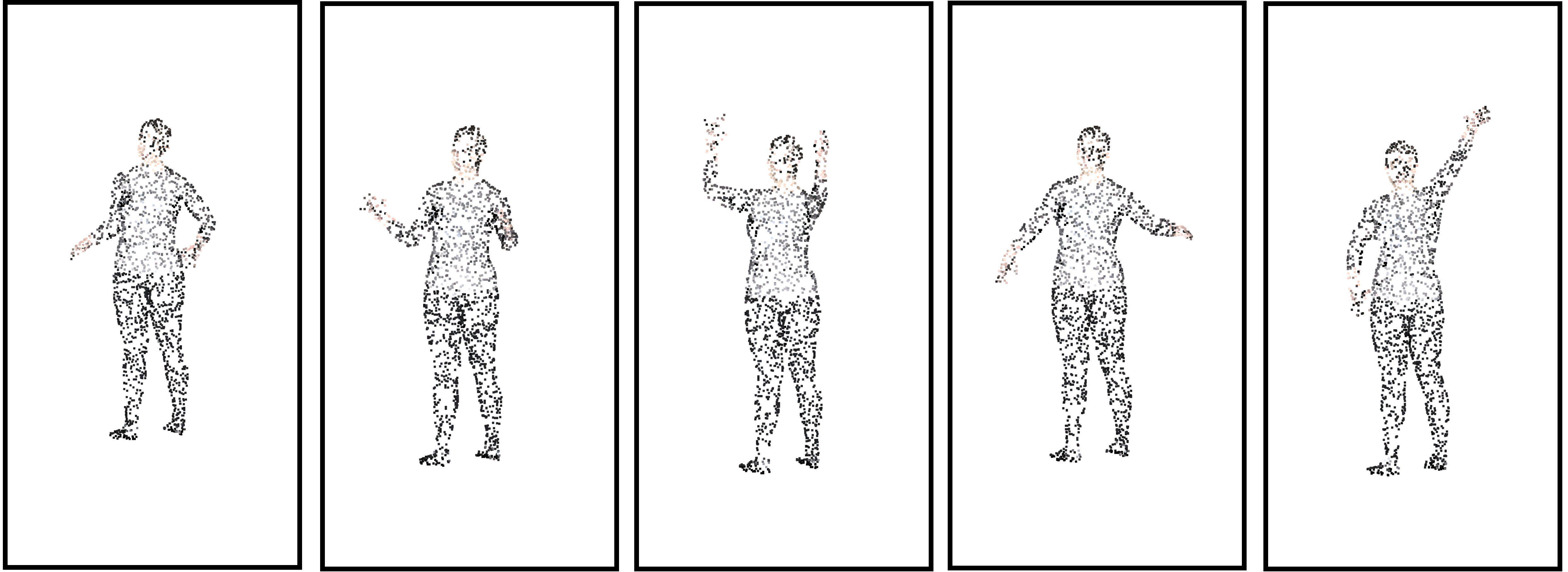}
    \end{minipage}
    
    \vspace{0.3cm} 
    
    \begin{minipage}{\linewidth}
        \centering
        \includegraphics[width=\linewidth]{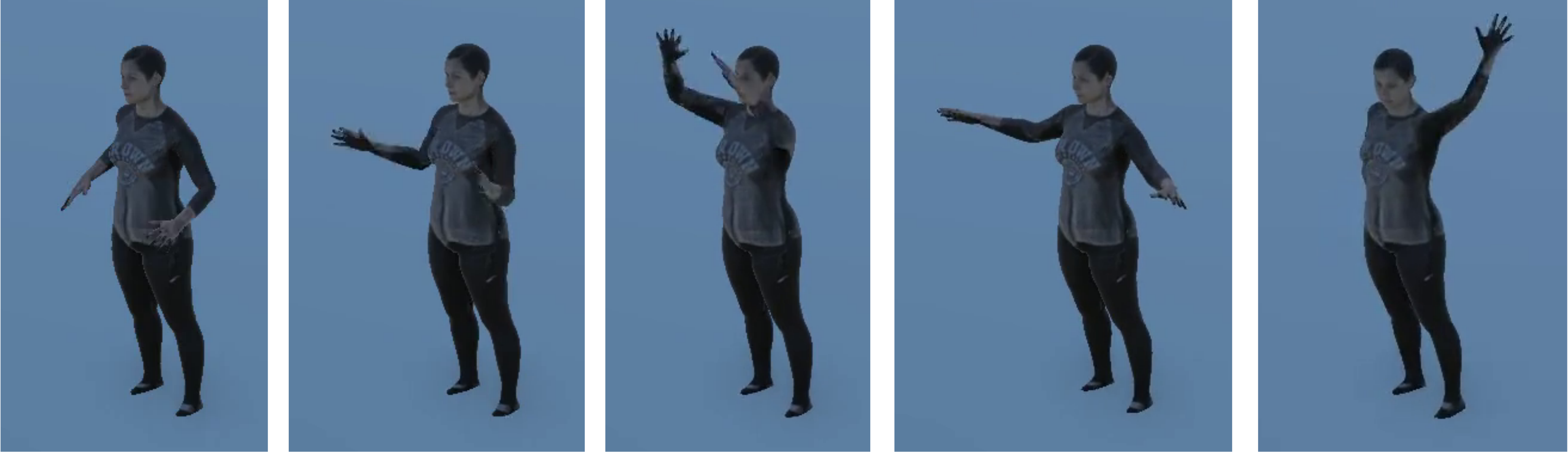}
    \end{minipage}
    
    \vspace{0.4cm} 
    
    \begin{minipage}{\linewidth}
        \small 
        \begin{itemize}
            \setlength{\itemsep}{4pt} 
            
            \item \textbf{Q:} What occurs immediately after the person forms a V-shape with their arms?\\
                  \textbf{GT:} The person moves the arms downward, resting the left hand on the hip while stretching the right arm out to the side.\\
                  \textbf{4DVLM:} The person performs a rapid, alternating arm swing to the right side.
                  
            \item \textbf{Q:} How are the person's hands and arms positioned after the initial shoulder-height lift?\\
                  \textbf{GT:} The person raises the arms into a broad V-shape with the palms of the hands directed forward.\\
                  \textbf{4DVLM:} The person's hands are held together in front of the chest with the palms facing upward.
                  
            \item \textbf{Q:} What is the position of the person's lower body throughout the activity?\\
                  \textbf{GT:} The person remains standing in a single spot with the feet firmly planted and the lower body completely still.\\
                  \textbf{4DVLM:} The person maintains a stable and stationary stance with their feet planted firmly on the ground.
                  
            \item \textbf{Q:} What does the person do with their left arm after extending it out to the side?\\
                  \textbf{GT:} The person pulls the left arm in toward the chest area and then moves it upward toward the face.\\
                  \textbf{4DVLM:} The person brings the left arm back toward the body and down toward the waist.
        \end{itemize}
    \end{minipage}
    
    \caption{Qualitative results for \texttt{sequence\_000048}. The top row visualizes the point cloud sequence, while the bottom row displays the corresponding rendered video. The corresponding question-answering pairs, ground truth (GT), and our model's predictions (4DVLM) are provided below.}
    \label{fig:seq_000048}
\end{figure}

\begin{figure}[!h]
    \centering
    
    \begin{minipage}{\linewidth}
        \centering
        \includegraphics[width=\linewidth]{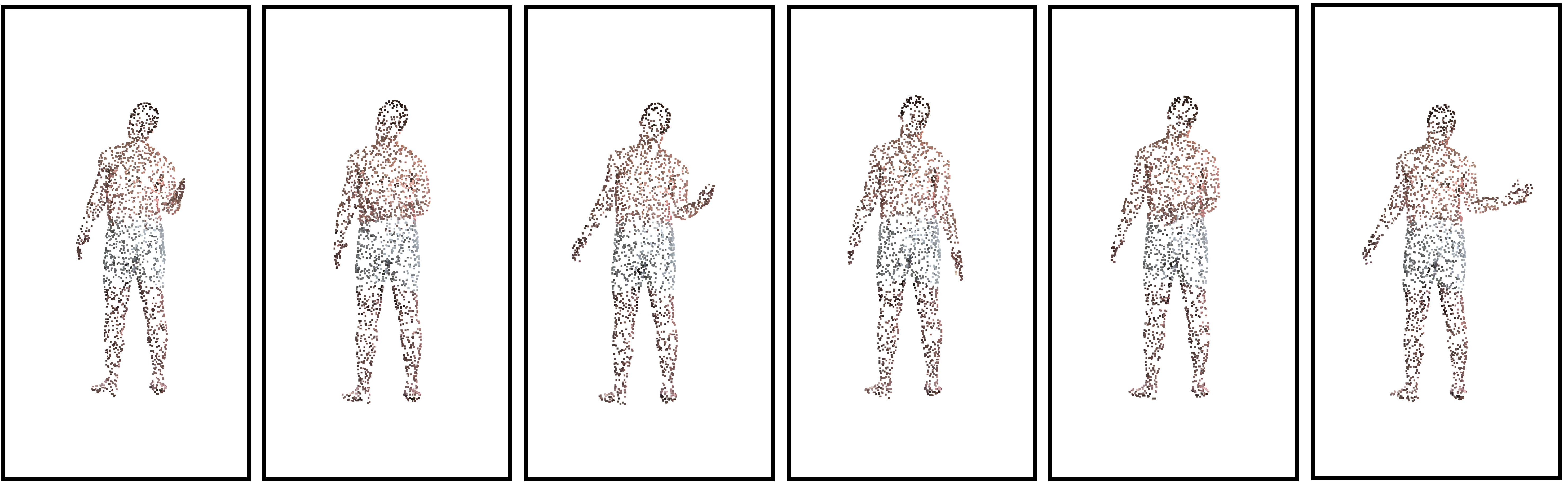}
    \end{minipage}
    
    \vspace{0.3cm} 
    
    \begin{minipage}{\linewidth}
        \centering
        \includegraphics[width=\linewidth]{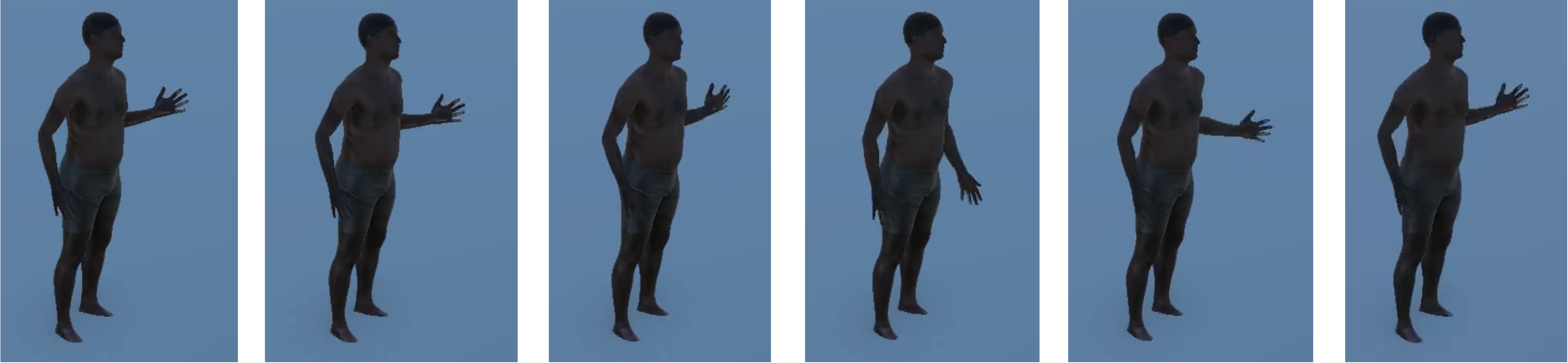}
    \end{minipage}
    
    \vspace{0.4cm} 
    
    \begin{minipage}{\linewidth}
        \small 
        \begin{itemize}
            \setlength{\itemsep}{4pt} 
            
            \item \textbf{Q:} What is the configuration of the right hand during the upward motion?\\
                  \textbf{GT:} The person keeps the right hand open with the fingers extended during the ascent.\\
                  \textbf{4DVLM:} The person keeps the right hand in a neutral position with the palm facing forward.
                  
            \item \textbf{Q:} How is the person's lower body positioned during the sequence?\\
                  \textbf{GT:} The person stands with straight legs placed about shoulder-width apart to maintain a stable and still base.\\
                  \textbf{4DVLM:} The person maintains a stable and upright posture with their feet planted firmly on the ground.
                  
            \item \textbf{Q:} What happens after the person raises their hand to mid-chest height?\\
                  \textbf{GT:} The person lowers the arm back down toward the hip before beginning the next repetition.\\
                  \textbf{4DVLM:} The person lowers their hand back down to their side.
                  
            \item \textbf{Q:} What movement does the person perform with their right arm?\\
                  \textbf{GT:} The person repeatedly lifts and lowers the right forearm by pivoting at the elbow in a rhythmic flexion and extension motion.\\
                  \textbf{4DVLM:} The person raises the right arm to a vertical position and then brings it down toward the left side of the body.
        \end{itemize}
    \end{minipage}
    
    \caption{Qualitative results for \texttt{sequence\_000055}. The top row visualizes the point cloud sequence, while the bottom row displays the corresponding rendered video. The corresponding question-answering pairs, ground truth (GT), and our model's predictions (4DVLM) are provided below.}
    \label{fig:seq_000055}
\end{figure}

\begin{figure}[!h]
    \centering
    
    \begin{minipage}{\linewidth}
        \centering
        \includegraphics[width=\linewidth]{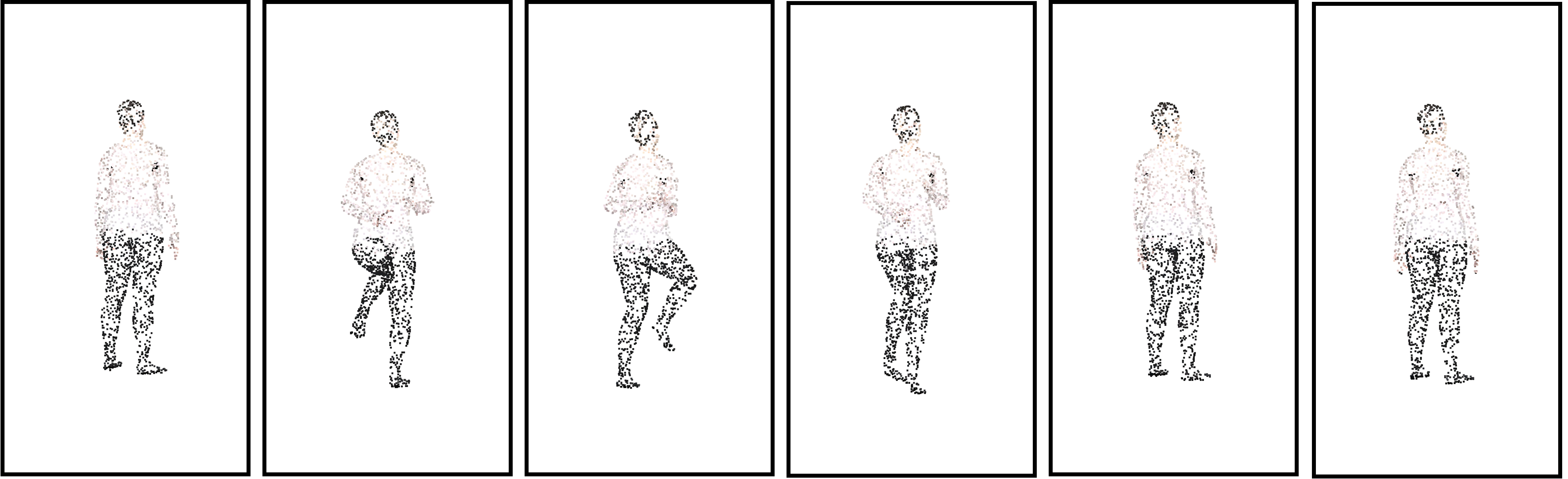}
    \end{minipage}
    
    \vspace{0.3cm} 
    
    \begin{minipage}{\linewidth}
        \centering
        \includegraphics[width=\linewidth]{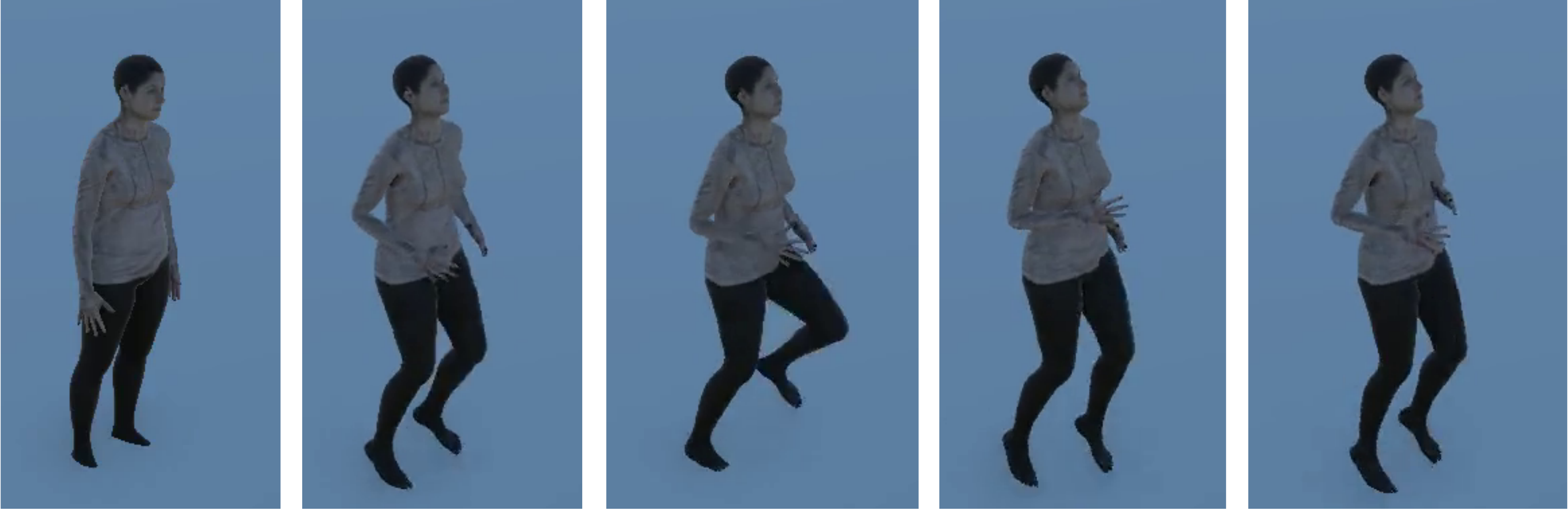}
    \end{minipage}
    
    \vspace{0.4cm} 
    
    \begin{minipage}{\linewidth}
        \small 
        \begin{itemize}
            \setlength{\itemsep}{4pt} 
            
            \item \textbf{Q:} What is the posture of the person's upper body during the jogging motion?\\
                  \textbf{GT:} The person maintains an upright torso throughout the duration of the stationary jog.\\
                  \textbf{4DVLM:} The person maintains a slight forward lean with the torso tilted slightly forward.
                  
            \item \textbf{Q:} How are the person's arms positioned throughout the main activity?\\
                  \textbf{GT:} The person keeps the elbows bent and swings the arms forward and backward in coordination with the opposite leg movements.\\
                  \textbf{4DVLM:} The person keeps their arms hanging loosely at their sides with the elbows bent.
                  
            \item \textbf{Q:} What is the final action performed in the sequence?\\
                  \textbf{GT:} The person slows their pace until both feet are lowered back to the ground and they return to a relaxed standing posture.\\
                  \textbf{4DVLM:} The person returns to a neutral standing position with both feet placed together.
                  
            \item \textbf{Q:} What does the person do once the initial movement is established?\\
                  \textbf{GT:} The person performs a rhythmic, alternating jogging motion in place, driving their knees upward while the arms swing in opposition to the legs.\\
                  \textbf{4DVLM:} The person performs a series of repetitive, rhythmic, and synchronized arm movements.
        \end{itemize}
    \end{minipage}
    
    \caption{Qualitative results for \texttt{sequence\_000099}. The top row visualizes the point cloud sequence, while the bottom row displays the corresponding rendered video. The corresponding question-answering pairs, ground truth (GT), and our model's predictions (4DVLM) are provided below.}
    \label{fig:seq_000099}
\end{figure}

Here we provide some figures of our \dataset{}-VQA segment samples in Fig.~\ref{fig:seq_000022} to Fig.~\ref{fig:seq_000099}. Each sample includes VQA samples, a point cloud sequence with rendered video sequence used for evaluations with video-LLM models.

%
%

\end{document}